\documentclass[final,3p,times,twocolumn]{elsarticle}

\usepackage{amssymb}
\usepackage{amsmath}

\usepackage{multirow}

\usepackage{url} 
\usepackage[dvipsnames]{xcolor}

\usepackage[table]{xcolor}

\begin{document}

\begin{frontmatter}



\title{Adaptive Image Zoom-in with Bounding Box Transformation for UAV Object Detection}


\author[1,2]{Tao Wang}



\ead{twangnh@gmail.com}



\author[1,2]{Chenyu Lin}
\ead{chenylin@scu.edu.cn}
\author[1,2]{Chenwei Tang}
\ead{tangchenwei@scu.edu.cn}
\author[1,2]{Jizhe Zhou}
\ead{jzzhou@scu.edu.cn}
\author[3]{Deng Xiong}
\ead{dxiong@stevens.edu}
\author[4]{Jianan Li}
\ead{lijianan@bit.edu.cn}
\author[5]{Jian Zhao}
\ead{zhaoj90@chinatelecom.cn}
\author[1,2]{Jiancheng Lv\corref{cor1}}
\ead{lvjiancheng@scu.edu.cn}
\cortext[cor1]{corresponding author}



\affiliation[1]{o={College of Computer Science, Sichuan University}, c={Chengdu}, p={610065}, cy={PR China}}
\affiliation[2]{o={Engineering Research Center of Machine Learning and Industry Intelligence, Ministry of Education}, c={Chengdu}, p={610065}, cy={PR China}}

\affiliation[3]{organization={Stevens Institute of Technology},
    city={Hoboken},
    postcode={NJ 07030}, 
    state={New Jersey},
    country={United States}}

\affiliation[4]{organization={Beijing Institute of Technology},
    city={Beijing},
    postcode={100053}, 
    country={P. R. China}}

\affiliation[5]{organization={China Telecom Institute of AI},
    city={Beijing},
    postcode={100053}, 
    country={P. R. China}}

\begin{abstract}
Detecting objects from UAV-captured images is challenging due to the small object size. In this work, a simple and efficient adaptive zoom-in framework is explored for object detection on UAV images. The main motivation is that the foreground objects are generally smaller and sparser than those in common scene images, which hinders the optimization of effective object detectors. We thus aim to zoom in adaptively on the objects to better capture object features for the detection task. To achieve the goal, two core designs are required: \textcolor{black}{i) How to conduct non-uniform zooming on each image efficiently? ii) How to enable object detection training and inference with the zoomed image space?}
Correspondingly, a lightweight offset prediction scheme coupled with a novel box-based zooming objective is introduced to learn non-uniform zooming on the input image. Based on the learned zooming transformation, a corner-aligned bounding box transformation method is proposed. The method warps the ground-truth bounding boxes to the zoomed space to learn object detection, and warps the predicted bounding boxes back to the original space during inference. 
We conduct extensive experiments on three representative UAV object detection datasets, including VisDrone, UAVDT, and SeaDronesSee.
The proposed ZoomDet is architecture-independent and can be applied to an arbitrary object detection architecture. 
Remarkably, on the SeaDronesSee dataset, ZoomDet offers more than 8.4 absolute gain of mAP with a Faster R-CNN model, with only about 3 ms additional latency. The code is available at \url{https://github.com/twangnh/zoomdet_code}. 
\end{abstract}

\begin{highlights}
\item Adaptive spatial transformation on images can magnify the details of the object efficiently.
\item Box transformation enables detector training and inference with image transformation.
\item Experiments on various UAV image datasets shows effective gains with small cost.
\item Flexible modular design enables integration with other approaches and task scenarios.
\end{highlights}


\end{frontmatter}





\section{Introduction}

Object detection on UAV (Unmanned Aerial Vehicle) captured images~\cite{xia2018dota,yang2019clustered,ding2021object} has been an active research topic due to its important role in many remote sensing applications, for example, UAV-aided environmental monitoring and disaster response systems.
Driven by the development of deep convolution neural networks (CNNs), UAV object detection has advanced significantly ~\cite{yang2019clustered,zhu2021detection,li2022oriented,li2020density,deng2020global,han2021align} in recent years. 
Since many objects in UAV images are small in size, distributed non-uniformly and sparsely due to high viewing attitude and irregular viewing angle~\cite{yang2019clustered}, one major research topic is studying how to detect these objects effectively.
\textcolor{black}{Current SOTA methods process images by uniformly cropping them into patches and applying detection independently to each patch}~\cite{koyun2022focus,duan2021coarse,wang2020object,xu2021adazoom,li2020density,yang2019clustered}. These patch-based methods offer stable and significant gains in detection accuracy. However, they only help address the cross-patch distribution variance, while small or sparse objects may still exist within the image patch. Moreover, patch-based methods introduce significant extra computation cost and latency due to their complicated procedure and multiple forward passes, which is expensive for edge computing devices such as those running on a UAV.

\begin{figure}[t]
	\centering
	\includegraphics[width=0.9\linewidth]{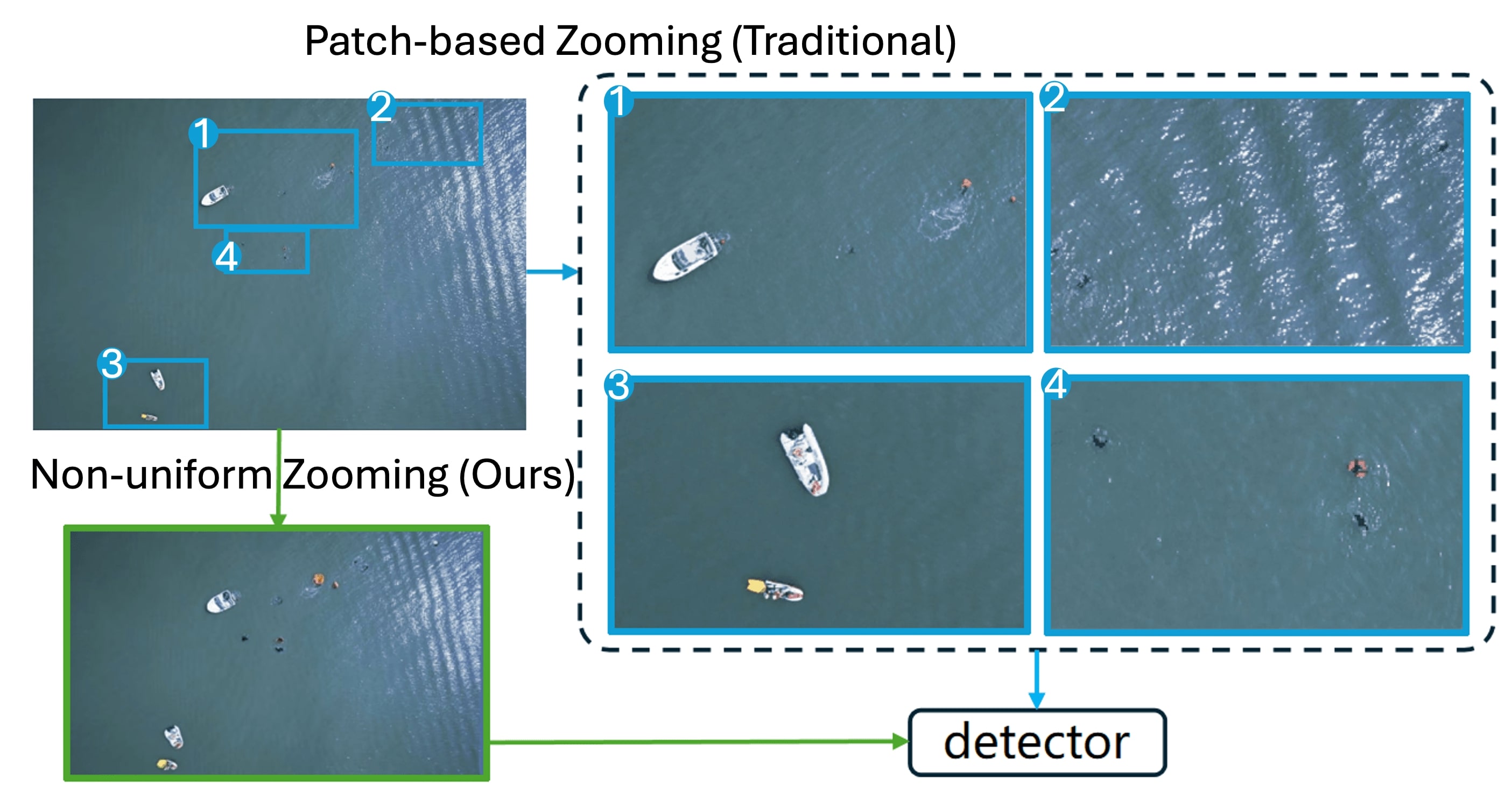}
	\caption{Illustration of Motivation. Unlike prior patch-based methods that generate patch candidates and perform object detection on the patches, the goal is to directly perform detection on an adaptively zoomed image.}
	\label{fig:idea}
\end{figure}

Inspired by previous work~\cite{recasens2018learning} that learns non-uniform image zooming to magnify important details and thus obtains improved image classification performance, \textcolor{black}{we aim to develop a non-uniform zooming framework that adaptively magnifies the object regions within the image for improving detection optimization, while maintaining the efficiency by only processing a single image.}
Fig.~\ref{fig:idea} shows the general idea of such non-uniform zooming for UAV object detection. The core design of the framework is the image transformation that maps discrete pixel coordinates in the output image space to the original image space. 
Such a framework is not only straightforward but also orthogonal to patch-based methods and can be combined to achieve optimal performance.

However, two unique challenges are identified when designing the adaptive zooming framework for object detection: \textbf{1)} The saliency-based method~\cite{recasens2018learning} computes the coordinate mapping with the average of neighboring coordinates weighted by image saliency and a predefined Gaussian distance kernel~\cite{recasens2018learning}. Such formulation is not only complex with the Gaussian kernel to tune, but also may cause heavy distortion to objects and the surrounding context due to the sharp peak nature of saliency prediction. \textbf{2)} Though the objects are magnified, the image transformation causes misalignment to ground-truth bounding boxes, which hinders the training and inference of detectors with zoomed images.

To address the above limitations and design an effective zooming framework, we introduce a simple offset-based object zooming coupled with a corner-aligned forward and backward bounding box transformation. Specifically, a lightweight offset predictor is used to regress spatial offsets that alter the original uniform grid coordinates, which can be used to instantiate the coordinate mapping of image transformation. This is partly motivated by the deformable convolution that predicts spatial offsets to apply the convolution operation~\cite{dai2017deformable,zhu2019deformable}, here, the offsets are predicted to apply an image transformation operation. To learn the offset prediction, a novel box-based zooming objective is developed, with which the ratio of zoomed bounding box area to the original one is maximized to learn object magnification. Unlike the saliency-based design in prior work~\cite{recasens2018learning}, the area under the bounding box is magnified and thus helps preserve the object and nearby context against extreme distortion. 

With the above offset prediction, input images can now be zoomed adaptively to focus on the objects, yet the original ground-truth bounding box annotations are invalid due to the spatial transformation. The gold transformation of bounding boxes to the zoomed space requires an accurate inverse of the coordinate mapping, which is difficult. We instead develop an approximated box transformation scheme. Specifically, the two corner points are utilized as a representation of bounding boxes, and their nearest mapped coordinates in the zoomed image space are found to obtain the transformed bounding boxes. This is achieved by treating the forward mapping computed in the above offset zooming process as a lookup table and performing a nearest neighbor search on the inverted table. During inference, the predicted bounding boxes are likewise transformed to the original image space for evaluation. Such a transformation is empirically validated to introduce small errors. This is performed by computing the IoU of the original bounding boxes and the ones that are forward transformed to the zoomed image space and then backward transformed to the original space.

The proposed image zooming and box transformation is integrated to obtain an efficient ``zoom in and detect" framework, which is named \textit{ZoomDet}. To validate the framework, thorough experiments are conducted with three representative UAV object detection datasets of urban and maritime scenarios: VisDrone~\cite{zhu2021detection}, UAVDT~\cite{du2018unmanned}, and SeaDroneSee~\cite{varga2022seadronessee}. For detection architecture, the two-stage Faster R-CNN detector~\cite{ren2015faster} and the popular one-stage YOLOv8 detector~\cite{Jocher_Ultralytics_YOLO_2023} are validated. ZoomDet brings significant gains to the baseline detector, with minor extra latency. For example, a Faster R-CNN model equipped with ZoomDet obtains a remarkable \textbf{8.4} mAP improvement on the SeadroneSee dataset, with only 2ms additional inference time. With similar latency overhead, ZoomDet introduces about 2.0 improvement in mAP on VisDrone and UAVDT datasets. 
Furthermore, ZoomDet shows orthogonal improvement to other SOTA zooming-based methods for UAV object detection, e.g., further improving the SOTA patch-based zooming and implicit zooming methods by more than 1.0 AP on small objects.
Additionally, we present qualitative examples showing that the zoomed image can help improve large aerial vision-language models~\cite{kuckreja2023geochat}, which shows the potential of further developing the proposed zoomed mechanism in more advanced aerial image understanding scenarios.

The contributions are summarized as follows:
\textcolor{black}{
\textbf{1)} An in-depth analysis of the background and challenges in designing a non-uniform zooming-based object detection framework with UAV images is presented, revealing the major difficulties of the coordinate mapping formulation and bounding box transformation between the original and zoomed image space.
\textbf{2)} A direct offset prediction mechanism supervised by a novel object zooming objective is introduced, which bridges the offset with object magnification and learn adaptive and accurate zooming on objects. Based on the learned non-uniform zooming, a corner-aligned bounding box transformation is developed to achieve flexible detector training and inference in the zoomed space.
\textbf{3)} For the first time, we show that a non-uniform image sampling approach can be applied to the challenging UAV object detection domain, bringing significant performance gain on various benchmark datasets, baselines, taking minor extra cost in parameters, computation, and latency. 
\textbf{4)} The proposed “zoom-and-detect” framework shows good generality and extensibility. It is compatible with various detection architectures (Faster R-CNN, YOLO). Additionally, it can be combined with other zooming methods, such as patch-based methods and feature-level enhancement, to obtain further gains. Furthermore, it can be extended to large aerial vision-language models on visual question answering, indicating its potential beyond pure detection tasks.
}

\section{Related Work}
\label{sec:related}

\subsection{Object Detection and its Development on UAV Images} 
Object detection is a fundamental task in the field of computer vision. Based on the development of deep learning frameworks~\cite{paszke2017automatic} and generations of GPU computing devices, deep convolution neural networks~\cite{lecun2015deep} have greatly revolutionized the object detection task. With pioneer models of the two-stage Faster R-CNN family~\cite{ren2015faster,girshick2015fast,he2017mask}, the latter works are mainly led by two directions, one-stage models and multi-stage cascaded models. The representative one-stage models are mainly the YOLO series~\cite{redmon2016you,redmon2017yolo9000,redmon2018yolov3,bochkovskiy2020yolov4,li2022yolov6,wang2023yolov7,Jocher_Ultralytics_YOLO_2023,varghese2024yolov8,wang2024yolov10} models and SSD series~\cite{liu2016ssd,zhang2018single,womg2018tiny,woo2018stairnet} models. Under the one-stage paradigm, the research community also explored other architecture variations, such as anchor-free frameworks~\cite{duan2019centernet,tian2019fcos,law2018cornernet,law2019cornernet}, where the object bounding boxes are directly regressed from the features, without the help of prior bounding boxes. The other branch of detection model development is the multi-stage cascaded architecture~\cite{cai2018cascade,chen2019hybrid}, which pursues accurate bounding box regression by cascading multistage prediction heads. 
Unlike the traditional one-stage and multi-stage models, in recent years, researchers have developed a fully end-to-end object detection architecture based on a transformer network. The transformer-based models~\cite{carion2020end,zhu2020deformable,wang2021pnp,roh2021sparse,zhang2022dino} require no post-processing as in traditional models and demonstrate SOTA performance.

The milestone models are mainly developed with common object detection datasets such as COCO~\cite{lin2014microsoft}. Benefiting from this development, many prior works~\cite{yang2019clustered,meethal2023cascaded} transfer the milestone models to UAV images and achieve good performance. However, UAV images are mainly captured in aerial view, thus the domain gap is large, \textit{e.g.}, the cars are much smaller, and the appearance differs from that of natural images. As a result, the research effort on UAV object detection mainly focuses on improving the milestone models with domain-specific designs. 
For example, a large series of works focus on patch-based designs to improve large resolution images~\cite{yang2019clustered,duan2021coarse,wang2020object,xu2021adazoom,li2020density,yang2019clustered,kouris2019informed,plastiras2018efficient,plastiras2019edgenet}. In addition to improving standard bounding box detection accuracy as in normal object detection setting, some works study how to predict rotated bounding boxes to better fit the aerial objects~\cite{Cheng2019LearningRA,qian2021learning,bi2025good}. Unlike those model-centric designs, others develop robust methods against data distribution changes~\cite{chen2025teaching,ma2024hierarchical}, application domain data~\cite{yan2025rf,tao2024small,li2025specdetr}, and model learning efficiency~\cite{zhang2025s3od,zhu2025tiny,li2025adaptive}.
This work is motivated by the size and spatial distribution challenge of aerial objects, and focuses on developing a general non-uniform image sampling approach to improve the UAV object detection accuracy.

\textcolor{black}{\subsection{Non-uniform Zooming-based Methods for Recognition and Detection}}
Adaptive image warp (transformation) is first explored in the seminal spatial transformer networks~\cite{jaderberg2015spatial}, where they predict the affine transformation parameters to transform an input image. Instead of a parametric affine transformation, latter works~\cite{recasens2018learning} explored a saliency-based pixel grid pulling formulation to predict the sampling grid; such a formulation helps magnify local regions and is demonstrated to offer high performance improvement to recognition tasks such as fine-grained image classification and gaze estimation. 
Some following works tried extending the saliency-based samping~\cite{recasens2018learning} to object detection~\cite{thavamani2021fovea, Thavamani_2023_CVPR}. \textbf{The major difference is three-fold}: 1) Due to the difficulty of computing inverse transformation, they are optimized in the original image space by either mapping the bounding boxes or image features back to the original space, which hinders learning more accurate detection on the zoomed image space. 2) They adopted the saliency-based parametrization of image transformation~\cite{recasens2018learning}, which is cumbersome and leads to heavy distortion of objects and surrounding context. 3) They only focus on the common scene images, while this work studies the challenging UAV image domain where objects are generally much smaller and distributed randomly in the image space (refer to the supplementary for an analysis of the data distribution). The learning signal is thus weakened and indirect for the zoomed object.
This work is related to these works, while simple offset-based transformation, along with bounding box transformation, are studied for training and inference with the zoomed image space.

\textcolor{black}{\subsection{Uniform Zooming-based Methods for Recognition and Detection}}
\textcolor{black}{
In addition to non-uniform zooming methods that explicitly perform non-uniform image zooming to magnify the object of interest, other works explore uniform zooming on the different levels, including full image, patches, or features, i.e., they keep the original uniform grid structure of the input image and perform global enhancement on the objects. These recent works can be categorized into patch-based zooming\cite{meethal2023cascaded,liu2024esod,li2024saccadedet,xie2025density}, uniform zooming\cite{liu2024small,nihal2024blurry,kim2024beyond,asif2025novel,zhang2023superyolo}, and implicit feature zooming\cite{wang2025learning,bian2025feature,bi2025towards,li2025remdet,wang2025detail,duan2025focal}. 1) Patch-based zooming: image patches are generated from the input image, then object detection is conducted on the uniformly enlarged patches to help better detect the smaller objects. The final detection result is usually obtained by combining detections on the crops\cite{xie2025density,liu2024esod}. 2) Uniform image zooming: the input images are super-resolved to a new version, with sharper details or reconstructed missing information. The new zoomed image is subsequently fed to object detection models. Recent method trains the super-resolution network with task-driven perceptual loss\cite{kim2024beyond}.
In addition, some works extend that to the feature level, by establishing a feature-to-image reconstruction pipeline\cite{zhang2023superyolo}, or feature-to-feature reconstruction\cite{liu2024small}. 3) Implicit feature zooming: instead of zooming on the image, this category mimics the human visual zooming by fusing multi-scale feature information, e.g., ``learning to zoom"\cite{wang2025learning} proposes a global contextual aggregation and a local feature enhancement scheme to improve small object feature responses. 
Similarly, other works develop within-scale detail feature enhancement modules and cross-scale feature fusion to merge multi-scale information and magnify small object signal\cite{wang2025detail,duan2025focal,li2025remdet,bi2025towards}.
}

\textcolor{black}{\subsection{Bounding Box Transformation}}
\textcolor{black}{
The above image zooming transformation is quite diverse and complex, which is driven by the inherent richness of image content. Unlike that, the bounding box representation is simpler, with a few coordinates. 
Thus, the transformation involving bounding boxes primarily focuses on adjusting the coordinate representation\cite{ren2016faster,cai2018cascade}. For example, Faster R-CNN transforms the initial coarse bounding box proposals into final detection boxes by learning offset regression networks\cite{ren2016faster}. Later, several researchers extend the transformation to a multiple-stage one and propose cascaded detection architectures\cite{cai2018cascade,chen2019hybrid}, or enhanced with reinforcement learning\cite{konig2019multi}. 
In addition to the regular bounding boxes, some works build a bounding box transformation scheme on oriented box detection scenarios\cite{ma2018arbitrary,qian2023building,kim2025nbbox}.
Kim et al.\cite{kim2025nbbox} develop basic transformation schemes such as scaling and rotation to augment training data. Qian et al\cite{qian2023building} propose a general transformation method between the regular bounding box regressing objective and the oriented case, which enables more effective oriented box regression training. 
Except for augmenting the training signal, other works explore angle regression transformation\cite{ma2018arbitrary} to improve precise oriented bounding box prediction. 
The above transformation preserves coordinate representation, while some research works expand the transformation to obtain other formats. e.g., 
Progressive deformation and regression are introduced to convert the original bounding box to multi-point polygons to capture object masks\cite{peng2020deep}.
Unlike current development that assumes a static input image, this work aims for a bounding box transformation that is associated with image transformation. The box transformation is associated with image transformation to be compatible with detector training and inference.
}

\subsection{Spatial Offset Learning with Neural Networks}
Image spatial offset prediction is a fundamental component in deep learning-based spatial regression tasks, enabling precise localization by refining coarse proposals or correcting discretization errors. In object detection, methods like Faster R-CNN~\cite{ren2015faster} and YOLO~\cite{redmon2017yolo9000,wang2024yolov10} regress bounding box offsets to adjust anchor boxes toward ground truth coordinates, often using normalized displacement and logarithmic scaling for stability. Similarly, keypoint estimation networks predict keypoint offsets to achieve fine-grained localization~\cite{shi2022end,zhang2021direct,law2018cornernet}. 
In addition to sparse offset prediction, pixel-wise offsets are employed in dense prediction tasks such as optical flow~\cite{wang2024sea} and instance segmentation~\cite{peng2020deep,liu2021dance} to align features or refine boundaries. 
Unlike the above offset prediction for explicit regression task prediction, several research efforts are related to implicit offset. For example, deformable convolutions (DCN) employ offset to dynamically adjust feature sampling locations~\cite{dai2017deformable,zhu2019deformable}. The approach is later explored in specific vision tasks such as object detection~\cite{zhu2020deformable} and pose estimation~\cite{chen2024meta} to improve feature extraction. This work explores adaptive offset prediction for parameterizing image transformation, which also belongs to implicit offset prediction, but is different from the deformable convolution network, as it does not involve distortion to images.

\section{Proposed Method}

We first introduce an overview of the zooming-based object detection approach and discuss the critical design choices, and then present detailed designs that instantiate an efficient zooming-based detection framework. Throughout the remaining sections, bold characters (\textit{e.g.}, $\textbf{x}$) are used to represent tensor or vector variables, and regular characters are used to represent scalar variables (\textit{e.g.}, $x$).
\subsection{Overview and Motivation}
\label{method_overview}
Given an input image $\textbf{I}$, the goal is to magnify the small objects by performing a non-uniform spatial transformation on the input image:
\begin{align}
    \textbf{I}'(x,y)= \textbf{I}(u,v)
\end{align}
where $\textbf{I}'$ denotes the zoomed version of the input image, $(x,y)$ is the discrete spatial pixel coordinates on the output image space (zoomed space), and $(u,v)$ is the continuous spatial pixel coordinates on the input image space (original space). Here, the pixel coordinates are assumed to have been normalized. The pixel values on the continuous spatial coordinates are obtained through bilinear interpolation. The transformation is based on a mapping $\mathcal{T}$ that maps the coordinates:
\begin{align}
    \mathcal{T}:(x,y)\rightarrow(u,v)
\end{align}
With proper mapping, the image transformation magnifies the objects of interest and thus helps better detect them. The challenging design choices to instantiate such a zooming-based object detection framework are:
\begin{itemize}
    \item \textbf{i)} \textit{The parametrization of the mapping $\mathcal{T}$ should sufficiently capture the location of objects and map densely on them to achieve effective magnification.} 
    \item \textbf{ii)} \textit{The transformation causes misalignment to the bounding box labels; thus, it is crucial to design a box label transformation method between the original and zoomed spaces, which enables effective training and inference procedures for object detection models.}
\end{itemize}

Previous work based on zooming~\cite{recasens2018learning} parameterizes the mapping $\mathcal{T}$ with a complex saliency-guided coordinate pulling mechanism to boost fine-grained classification and gaze estimation tasks. Specifically, given a saliency map $\textbf{S}$ that is predicted by a neural network and captures the task-related details such as object discriminative parts~\cite{recasens2018learning}, a weighted average is conducted with a Gaussian distance kernel $\textbf{k}((x, y),(x', y'))$ to obtain the mapped spatial coordinates:
\begin{align}
    \mathcal{T}_u(x, y) =
    \frac{\sum_{x',y'}\textbf{S}(x', y')\textbf{k}((x, y),(x', y'))x'}{\sum_{x',y'}\textbf{S}(x', y')\textbf{k}((x, y),(x', y'))}
    \label{eq3}
\end{align}
\begin{align}
    \mathcal{T}_v(x, y) =
    \frac{\sum_{x',y'}\textbf{S}(x', y')\textbf{k}((x, y),(x', y'))y'}{\sum_{x',y'}\textbf{S}(x', y')\textbf{k}((x, y),(x', y'))}
    \label{eq4}
\end{align}
where $(x,y)$ defines the center of Gaussian kernel and $(x', y')$ iterates over neighboring spatial locations.
The weighting of the saliency map pulls transformation coordinates to salient image regions, and the Gaussian kernel reduces the weights for non-center locations and thus avoids the cases where near coordinates collapse to the same spatial location.

\textbf{Limitations of Saliency-based Parametrization}
Though effective for image-level understanding tasks like classification and gaze estimation, the saliency-based parametrization causes severe distortion to objects and the surrounding background, which could harm instance-level object detection. This is due to the fact that saliency tends to peak on certain spatial regions and thus causes excessive sampling on nearby locations (with equations~\ref{eq3} and ~\ref{eq4}).
Furthermore, the non-uniform zooming causes bounding box misalignment, which is difficult to recover as the inverse mapping 
$\mathcal{T}^{-1}$ is difficult to solve; thus, it is infeasible to transform the bounding box label to the zoomed space. 
\textit{These limitations hindered the application of the zooming to UAV object detection domain, where objects are generally small in scale and thus sensitive to image distortion and box label variance.}

Based on the above observation, in this work, the complex saliency-based parametrization of $\mathcal{T}$ is abandoned, and an offset-based parametrization with a novel bounding box label transformation scheme is introduced. The offset-based parametrization learns to predict the spatial offsets of coordinates, with a box-based zooming objective that induces less distortion. 
The bounding box transformation scheme efficiently converts the ground-truth bounding box to zoomed space during training and converts the prediction to original space during inference.

\subsection{Offset-based Mapping Parametrization}
To predict the transformation mapping $\mathcal{T}$, a small convolution neural network $f_\theta$ is employed to predict spatial offsets ($\Delta x$, $\Delta y$) for each pixel location, where $\theta$ denotes the learnable parameters. $f_\theta$ is named as OffsetNet. The predicted offsets are added to the pixel coordinates $(x,y)$, thus the mapping is simply parametrized as:
\begin{align}
    \mathcal{T}_u(x, y) = x + \Delta x
\end{align}
\begin{align}
    \mathcal{T}_v(x, y) = y + \Delta y
\end{align}
Unlike saliency-based parametrization, which tends to focus on the center region of the object and causes heavy distortion, \textit{this offset-based zooming parametrization equally weights the object region and mitigates the distortion}. Additionally, the simple weight decay term ($\frac{1}{2}||\theta^{2}_{2}||$) can be used to regularize the transformation, to enforce uniform image sampling, and thus help mitigate the distortion. 

To save computation, the input image is downsampled to obtain a smaller version $\textbf{I}_d$, and used to predict the offsets on $\textbf{I}_d$ with $f_\theta$:
\begin{align}
    \boldsymbol{\Delta}_d = f_\theta(\textbf{I}_d)
\end{align}
the output offsets $\boldsymbol{\Delta}_d$ are then interpolated to the zoomed image size to obtain $\boldsymbol{\Delta}$, which is two-dimensional and corresponds to $\Delta x$ and $\Delta y$. 

Simply learning $f_\theta$ implicitly with the gradients back-propagated from object detection loss is insufficient, as the detection loss is indirect for object magnification.
As a result, it is necessary to design effective learning objectives.
\textit{However, unlike saliency prediction, which may be easily learned from object bounding box annotations, it is challenging to define offset prediction objectives, as sampling offset is not directly related to the magnification of object regions.}

\subsection{Object Zooming Loss}
\label{method_obj_zooming_loss}

To address the challenge, an \textit{object zooming loss} is introduced to learn the mapping offset prediction. Concretely, for each bounding box $\textbf{b$_i$}$ on the input image, its box area is treated as a mask $\textbf{a$_i$}$ on the input image space, then the same image transformation with mapping $\mathcal{T}$ for the input image is conducted on the mask to obtain a zoomed mask $\textbf{a$_i$}'$: $\textbf{a$_i$}'(x,y)=\textbf{a$_i$}(u,v)$. A zooming ratio can be computed as:
\begin{align}
    m_i=sum(\textbf{a$_i$}')/sum(\textbf{a$_i$})
\end{align}
where $sum$ denotes the element sum operation for the mask, i.e., $sum(\textbf{a})=\sum_{h,w}\textbf{a}_{hw}$ with $h$ and $w$ as the spatial mask pixel index. This zooming ratio helps normalize the magnification of different objects. To learn the magnification, the Logarithm function is leveraged to design a loss function that maximizes the zooming ratio $m_i$. The intuition is that when $m_i$ is small, the loss should be larger to enforce the zooming. The loss function is designed as:
\begin{align}
    \mathcal{L}_{zoom} = \sum_{i}^{N} max\left( \log\left( \frac{\alpha+\epsilon}{m_i + \epsilon} \right),0\right)^{\beta}
    \label{eq:magloss}
\end{align}
where $\epsilon$ is a small constant that avoids infinite loss value. $\alpha$, $\beta$ are two hyperparameters controlling the loss shape. 
Specifically, $\alpha$ controls the magnification threshold, i.e., when $m_i>=\alpha$, the loss is zero, and no more magnification is enforced. $\beta$ adjusts the rate at which large magnification samples are down-weighted, which is similar to the exponent hyper-parameter used in Focal Loss~\cite{lin2017focal} and helps down-weight easy samples. A $max$ operation is added to clamp the value, thus avoiding negative loss.

The gradients are back-propagated through the bilinear interpolation operation of \textbf{a$_i'$} to the offset $\Delta x$ and $\Delta y$:

\begin{equation}
    \frac{\partial \mathcal{L}_{zoom}}{\partial \Delta x} = \sum_{x,y} \frac{\partial \mathcal{L}_{zoom}}{\partial \textbf{a}_i'(x,y)} \cdot \frac{\partial \textbf{a}_i'(x,y)}{\partial u} \cdot \frac{\partial u}{\partial \Delta x}
\end{equation}

\begin{equation}
    \frac{\partial \mathcal{L}_{zoom}}{\partial \Delta y} = \sum_{x,y} \frac{\partial \mathcal{L}_{zoom}}{\partial \textbf{a}_i'(x,y)} \cdot \frac{\partial \textbf{a}_i'(x,y)}{\partial v} \cdot \frac{\partial v}{\partial \Delta y}.
\end{equation}

where \textbf{a$_i'$} is obtained through bilinear interpolation of grid mask values of \textbf{a$_i$}:

\begin{equation}
    \begin{split}
        \mathbf{a}_i'(x, y) &= (1 - s)(1 - t)\mathbf{a}_i(h_1, w_1) + s(1 - t)\mathbf{a}_i(h_1, w_2) \\
        &\quad + (1 - s)t \mathbf{a}_i(h_2, w_1) + s t \mathbf{a}_i(h_2, w_2),
    \end{split}
\end{equation}
where $(h_1, w_1), (h_1, w_2), (h_2, w_1), (h_2, w_2)$ are four neighboring integer coordinates. And $s = u - w_1$, $t = v - h_1$ represent the fractional parts of $u$ and $v$.
And the gradient with respect to the bilinear interpolation operation can be derived as:
\begin{equation}
    \begin{split}
        \frac{\partial \mathbf{a}_i'(x,y)}{\partial u} &= (1-t)\left(\mathbf{a}_i(h_1,w_2) - \mathbf{a}_i(h_1,w_1)\right) \\
        &\quad + t\left(\mathbf{a}_i(h_2,w_2) - \mathbf{a}_i(h_2,w_1)\right)
    \end{split}
\end{equation}

\begin{equation}
    \begin{split}
        \frac{\partial \mathbf{a}_i'(x,y)}{\partial v} &= (1-s)\left(\mathbf{a}_i(h_2,w_1) - \mathbf{a}_i(h_1,w_1)\right) \\
        &\quad + s\left(\mathbf{a}_i(h_2,w_2) - \mathbf{a}_i(h_1,w_2)\right)
    \end{split}
\end{equation}

\begin{figure*}[h]
	\centering
	\includegraphics[width=\linewidth]{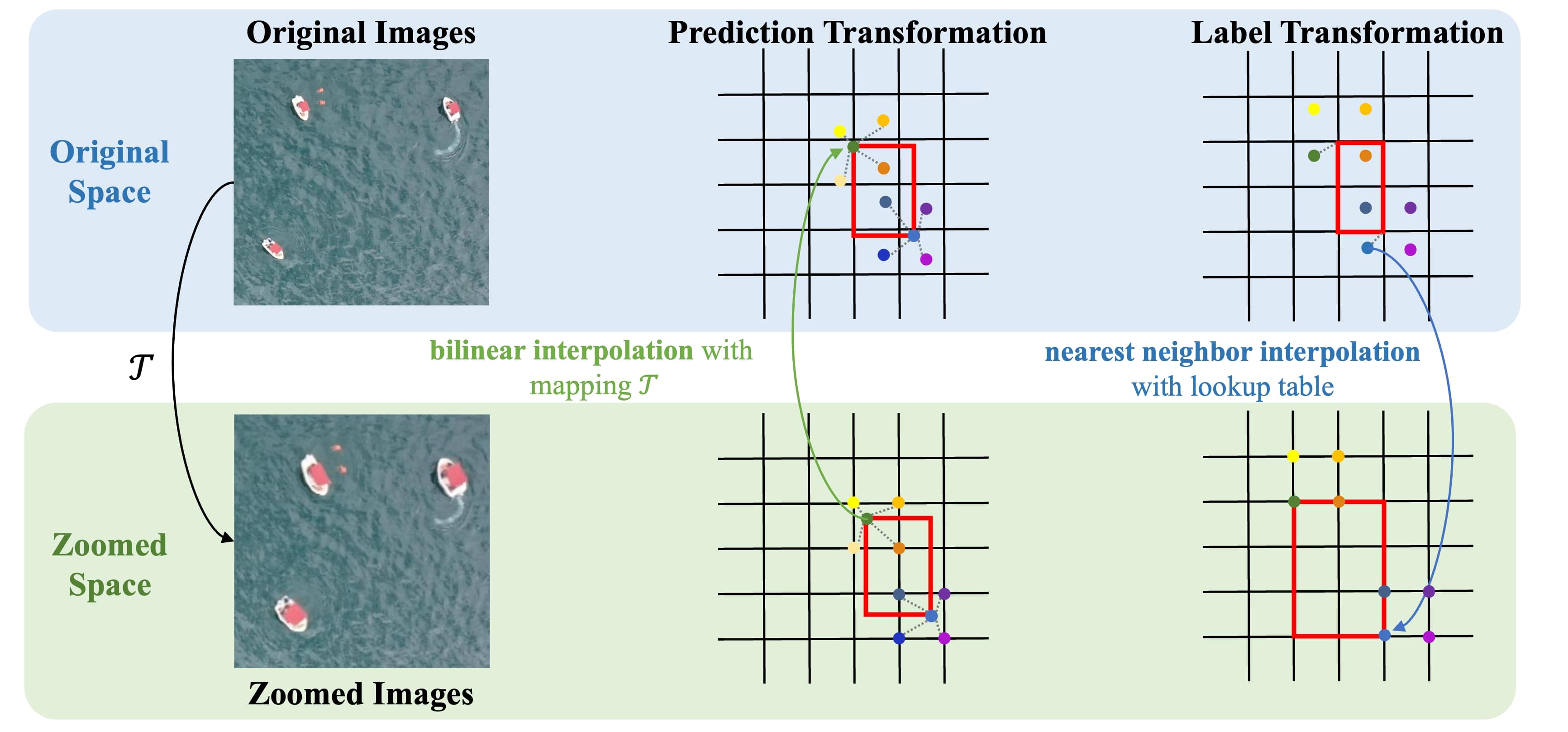}
	\caption{Illustration of forward bounding box label transformation (green arrow) and backward prediction transformation (blue arrow).}
	\label{fig:label_transform}
\end{figure*}

\subsection{Corner-aligned Box Transformation}
With the above offset-based zooming, the objects are enlarged, but the original bounding box annotations are invalid. Thus, the original boxes need to be transformed to the zoomed image space. However, it is challenging to solve the inverse mapping $\mathcal{T}^{-1}$, as $\mathcal{T}$ is non-uniform and data-dependent.
We thus propose a simple and efficient corner-aligned transformation to ``convert" the bounding box to the zoomed image space, without solving the inverse $\mathcal{T}^{-1}$. 

Specifically, given a bounding box $\textbf{b}$ in the original image space, we aim to map its higher-left corner $\textbf{c}_1=(u_1, v_1)$ and lower right corner $\textbf{c}_2=(u_2, v_2)$ to the zoomed space and use the mapped corners to instantiate a new bounding box. This corner-based representation has been widely used in the object detection community~\cite{law2018cornernet}. 
To achieve the transformation, the mapping computed in the above offset-based zooming process is treated as a lookup table:
\begin{align}
<\Phi:(x_j, y_j)\rightarrow \Psi:(u_j, v_j)>
\end{align}
and invert it as $<\Psi \rightarrow \Phi>$.
Then the nearest spatial neighbor of $\textbf{c}_1$ and $\textbf{c}_2$ are searched in $\Psi$, and used with the inverse lookup table to obtain the mapped corners $\textbf{c}'_1=(x_1,y_1)$ and $\textbf{c}'_2=(x_2,y_2)$:
\begin{align}
    \textbf{c}'_1, \textbf{c}'_2=\Psi(\mathcal{N}(\textbf{c}_1)), \Psi(\mathcal{N}(\textbf{c}_2))
\end{align}
where $\mathcal{N}$ denotes the nearest neighbor searching operation
\begin{equation}
\mathcal{N}(\mathbf{c}) = arg\ min_{(u_j, v_j) \in \Psi} \| (u_j, v_j) - \mathbf{c} \|_2
\end{equation}

The transformed bounding box is hence $\textbf{b}'=(\textbf{c}'_1,\textbf{c}'_2)$. With the proposed box transformation, the ground-truth bounding box is now aligned with the zoomed object, and object detector training can be conducted as usual. During inference, the predicted bounding boxes are transformed back to the original image space in a similar way. Since $\Phi$ is in a uniform pixel grid space, the nearest neighbor search is saved, and a bilinear interpolation is employed to obtain the transformed coordinates during inference. Fig.~\ref{fig:label_transform} shows examples with the spatial grid.






\subsection{Box Transformation Error Analysis}
\label{subsec:iou-bound}

To rigorously analyze the error induced by the above corner-aligned bounding box transformation, we derive a lower bound for the Intersection-over-Union (IoU) between the original ground-truth box $\mathbf{b}$ and the backward-transformed box $\hat{\mathbf{b}}$. 
The displacement between the two could arise from three aspects:
\begin{itemize}
    \item \textit{Forward Transformation}: Nearest-neighbor quantization during coordinate mapping, which could lead to errors in detector training.
    \item \textit{Detector Variance}: Localization uncertainty in bounding box predictions. The trained detector could have prediction variance for the bounding boxes.
    \item \textit{Backward Interpolation}: Bilinear interpolation during inverse box transformation could lead to additional errors.
\end{itemize}
 
For simplicity, the maximum displacement distance of corner points is denoted as $\tau$. Let the original bounding box have width $w$ and height $h$. The worst-case IoU occurs when corners are displaced along the same direction $\theta$ that leads to a reduction of intersection area and an increase of union area. 

The intersection over union is:
\begin{equation}
\label{eq:iou}
\text{IoU} = \frac{(w - \tau\cos\theta)(h - \tau\sin\theta)}{2wh - (w - \tau\cos\theta)(h - \tau\sin\theta)}
\end{equation}
by substituting $wsin\theta+hcos\theta$, the above can be proved to have the minimum value when $\theta=arctan(w/h)$:
\begin{equation}
\text{IoU} \geq \frac{wh - \tau\sqrt{w^2 + h^2}}{wh + \tau\sqrt{w^2 + h^2}}
\end{equation}
For $\tau \ll w,h$, the below result can be obtained:
\begin{equation}
\text{IoU}_{\text{min}} \approx 1 - 2\tau\sqrt{\frac{1}{w^2} + \frac{1}{h^2}}
\end{equation}

\noindent\textbf{Practical Implications:}
This bound implies the sensitivity of small objects to transformation errors, i.e., the smaller the size of the bounding boxes, the higher the IoU bound is. 
The IOU error is empirically validated in the later experiments.

\begin{figure*}[t]
	\centering
	\includegraphics[width=\linewidth]{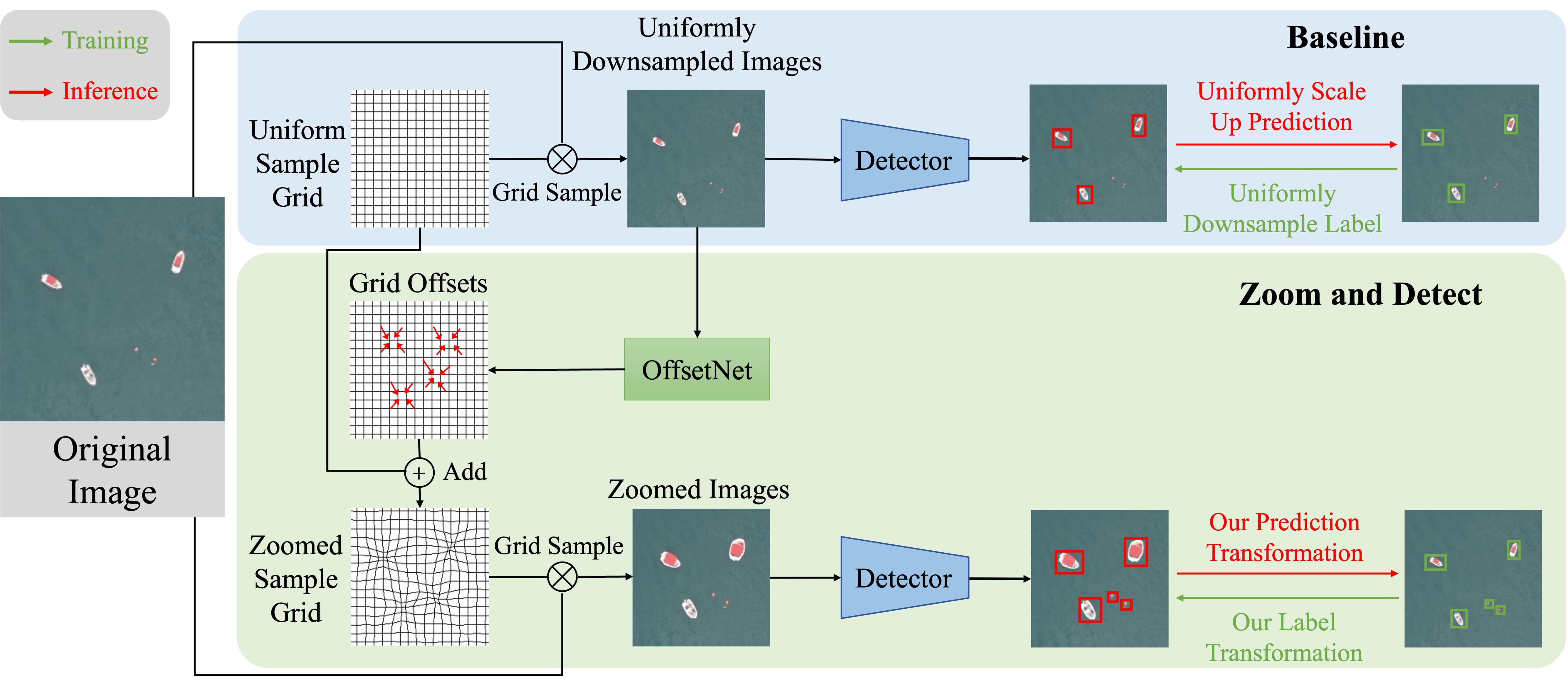}
	\caption{Overall framework of our proposed Method. Comparing the baseline approach that uniformly scales the image and performs detection, a non-uniform zoom-in is conducted to magnify the objects of interest and thus help better detect them; the detected bounding boxes are transformed back to obtain the final detection results.
    Note that the original input image may be large in resolution; thus, the baseline object detection pipeline involves a downsampling step to reduce resolution.}
	\label{fig:framework}
\end{figure*}

\subsection{Network Optimization}
Combining the above object zooming and bounding box transformation, an efficient object detection framework is obtained, which is named ZoomDet. The overall framework is shown in Fig.~\ref{fig:framework}. ZoomDet is \textbf{detector-agnostic} and can be applied to arbitrary object detection models. The whole framework is optimized with the object detection loss (usually defined with object classification and bounding box regression objectives) and the proposed object zooming loss:

\begin{align}
    \mathcal{L}=\mathcal{L}_{detection}+\mathcal{L}_{zoom}
\end{align}

\section{Experiments}
\textcolor{black}{In this section, thorough experiments and analyses with several representative datasets are conducted. The goal is to examine the following aspects:
1) The improvements brought by ZoomDet across datasets and object detection architectures, and comparison with close non-uniform zooming-based detection methods.
2) Comparison with other SOTA uniform zooming-based methods and Compatibility when integrated with these methods, such as patch-based zooming methods and implicit feature zooming methods.
3) Empirical analyses on the developed object zooming components, including the offset prediction network, the box-based zooming objective, and the bounding box transformation. 
4) Qualitative visualization and discussion on model prediction, cross-dataset box distribution, and extended application to other task scenarios beyond detection.
Unless otherwise noted in the later experiments, standard object detection training and inference setting is used; ad hoc techniques such as multi-scale training and inference are not used.}


\textcolor{black}{\subsection{Experimental Setup}}
\subsubsection{Datasets}
We use three publicly released UAV object detection benchmark datasets to conduct the experiments, including VisDrone~\cite{zhu2021detection}, UAVDT~\cite{du2018unmanned}, and SeaDronesSee~\cite{varga2022seadronessee}:

\paragraph{VisDrone} The VisDrone~\cite{zhu2021detection} dataset is an UAV object detection and tracking dataset collected in several urban/suburban street scenes. The VisDrone-2019 version is used, which consists of 10,209 images (6,471 for training, 548 for validation, and 3,190 for testing). The dataset is annotated with 10 common street object classes such as \textit{pedestrian} and \textit{car}. The image resolution of
the dataset is about 2, 000×1, 500 pixels. Since the test server is closed, the validation set is used to evaluate the proposed method following prior works~\cite{yang2019clustered}.

\paragraph{UAVDT} The UAVDT~\cite{du2018unmanned} dataset is a diverse UAV object detection dataset collected in urban scenes, capturing different views, times, and altitudes. The Benchmark-M set is used, which contains 24143 training images and 15,069 test images. The resolution of the image is about 1, 080 × 540 pixels. The dataset annotates 3 object categories: car, bus, and truck.

\paragraph{SeaDronesSee} The SeaDronesSee~\cite{varga2022seadronessee} dataset is collected in sea areas with aerial drone views, it aims to bridge the gap between land-based datasets and maritime-based ones. The dataset captures diverse viewing angles and attitudes. The object detection set is used, which consists of 8930 training images, 1547 validation images, and 3750 test images. The four common maritime UAV object categories are used: swimmer, boat, jetski, life\_saving\_appliances, and buoy. Since the test annotation is not released, the validation set is used to evaluate our method.

\subsubsection{Implementation Details}
 We implement ZoomDet based on mmdetection~\cite{mmdetection}, which is a widely used codebase for visual object perception tasks such as object detection and instance segmentation. To instantiate the offset predictor, we remove the layers after the second block of a ResNet-18 network (corresponding to conv\_4, conv\_5 in the ResNet paper) and append a single-layer convolution to predict two-dimensional outputs. The total number of parameters is 683,458 (11 convolution layers, 10 for ResNet, and 1 for the offset prediction). Such a convolution network is lightweight and achieves efficient offset prediction.
\textcolor{black}{Note that additional explicit regularization is not applied; the weight decay term is applied to regularize the offset predictor, which predicts zero offsets when the weights are regularized to zero, and thus regularizes the sampling to be uniform and preserve the original content. }

The Nearest neighbor function implemented in the Pytorch library (``torch-cluster-nearest-cuda.nearest") is employed to implement the box transformation. Unless otherwise noted, all models are trained for 24 epochs, using an SGD optimizer with a learning rate of 0.01, momentum of 0.9, and weight decay of 0.0001. All the models are trained using four Nvidia-4090 GPU cards. During training and inference, the image is resized with a window of $800 \times 1333$ or $800 \times 800$ for Faster R-CNN and YOLO models, respectively.

\begin{table*}[h!]
\centering
\caption{Results with \textbf{Faster R-CNN} (\textbf{revised in the response to add computation cost metric}). \textit{uniform} denotes the uniform downsampling baseline, \textit{saliency} means replacing proposed offset-based zooming parametrization with saliency parametrization used in prior works ~\cite{recasens2018learning}, while employing the proposed box transformation for detector training and inference. $ZR$ is a metric evaluating how much the objects are enlarged.}
\setlength{\abovecaptionskip}{5pt}
\renewcommand{\arraystretch}{1.4}
\setlength{\tabcolsep}{1.3mm}{
\begin{tabular}{c|c|ccccccccc}
   \hline
    \textbf{Dataset} & \textbf{Method} & $AP$ & $AP_{50}$ & $AP_{s}$ & $AP_{m}$ & $AP_{l}$ & $ZR_{s}$ & $ZR_{m}$ &
    $ZR_{l}$ & \textcolor{black}{FLOPs}\\
   \hline
   \multirow{5}*{VisDrone} & Uniform & 20.8 & 37.0 & 11.2 & 33.4 & \textbf{39.7} & 1.00 & 1.00 & 1.00 & \textcolor{black}{158.6} \\
    \cline{2-11}
    & Saliency~\cite{recasens2018learning} & 21.0 & 37.6 & 12.3 & 32.1 & 36.3 & 1.07 & 1.09 & 1.05 & \textcolor{black}{165.1}\\
    \cline{2-11}
    & FOVEA~\cite{thavamani2021fovea} & 21.5 & 37.9 & 12.9 & 32.8 & 36.9 & 1.05 & 1.07 & \textbf{1.07} & \textcolor{black}{166.7}\\
    \cline{2-11}
    & LZU~\cite{Thavamani_2023_CVPR} & 21.3 & 38.2 & 11.9 & 33.5 & 39.5 & - & - & - & \textcolor{black}{165.6}\\
    \cline{2-11}
   & ZoomDet & \textbf{22.8} & \textbf{39.9} & \textbf{13.9} & \textbf{34.4} & 38.1 & \textbf{1.17} & \textbf{1.10} & 1.04 & \textcolor{black}{163.4}\\
    \cline{2-11}
   \hline
    \multirow{5}*{UAVDT} & Uniform & 16.2 & 27.8 & 10.8 & 28.1 & 23.4 & 1.00 & 1.00 & 1.00 & \textcolor{black}{162.5}\\
    \cline{2-11}
   & Saliency~\cite{recasens2018learning} & 16.9 & 30.0 & 10.9 & 28.5 & 23.5 & 1.30 & \textbf{1.18} & 1.10 & \textcolor{black}{169.3}\\
   \cline{2-11}
   & FOVEA~\cite{thavamani2021fovea} & 16.7 & 30.5 & 11.0 & 28.9 & 22.1 & 1.37 & 1.15 & 1.02 & \textcolor{black}{170.4}\\
   \cline{2-11}
   & LZU~\cite{Thavamani_2023_CVPR} & 17.0 & 30.9 & 10.9 & 29.1 & 23.4 & - & - & - & \textcolor{black}{170.7}\\
    \cline{2-11}
   & ZoomDet & \textbf{18.0} & \textbf{32.7} & \textbf{11.3} & \textbf{30.2} & \textbf{23.5 }& \textbf{1.45} & 1.13 & \textbf{1.05} & \textcolor{black}{167.8}\\
      \hline
    \multirow{5}*{SeaDronesSee} & Uniform & 34.9 & 56.2 & 11.2 & 41.9 & 60.0 & 1.00 & 1.00 & 1.00 & \textcolor{black}{157.4}\\
    \cline{2-11}
   & Saliency~\cite{recasens2018learning} & 36.7 & 66.5 & 16.7 & 43.4 & 59.4 & 1.70 & 1.25 & 1.03 & \textcolor{black}{163.4}\\
    \cline{2-11}
   & FOVEA~\cite{thavamani2021fovea} & 39.9 & 71.8 & 22.7 & 44.0 & 59.8 & 2.32 & 1.19 & 1.02 & \textcolor{black}{164.0}\\
   \cline{2-11}
   & LZU~\cite{Thavamani_2023_CVPR} & 35.9 & 60.8 & 13.8 & 42.5 & 59.9 & - & - & - & \textcolor{black}{164.5}\\
    \cline{2-11}
   & ZoomDet & \textbf{43.3} & \textbf{79.7} & \textbf{34.3} & \textbf{44.5} & \textbf{60.1} & \textbf{2.67} & \textbf{1.34} & \textbf{1.03} & \textcolor{black}{163.3}\\
   \hline
\end{tabular}}
\label{tab:tab1}
\end{table*}

\begin{table*}[h!]
\centering
\caption{Results with \textbf{YOLOv8} (\textbf{revised in the response to add computation cost metric}). \textit{uniform} denotes the uniform downsampling baseline, \textit{saliency} means replacing proposed offset-based zooming parametrization with saliency parametrization used in prior works ~\cite{recasens2018learning}, while employ the proposed box transformation for detector training and inference. $ZR$ is a metric evaluating how much the objects are enlarged.}
\setlength{\abovecaptionskip}{5pt}
\renewcommand{\arraystretch}{1.4}
\setlength{\tabcolsep}{1.3mm}{
\begin{tabular}{c|c|ccccccccc}
   \hline
    \textbf{Dataset} & \textbf{Method} & $AP$ & $AP_{50}$ & $AP_{s}$ & $AP_{m}$ & $AP_{l}$ & $ZR_{s}$ & $ZR_{m}$ &
    $ZR_{l}$ & \textcolor{black}{FLOPs}\\
   \hline
   \multirow{5}*{VisDrone} & Uniform & 24.6 & 42.6 & 13.7 & 38.3 & 58.2 & 1.00 & 1.00 & 1.00 & \textcolor{black}{43.7}\\
    \cline{2-11}
    & Saliency~\cite{recasens2018learning} & 25.1 & 43.4 & 14.2 & 38.5 & 56.9 & 1.20 & 1.17 & \textbf{1.11} & \textcolor{black}{48.1}\\
    \cline{2-11}
    & FOVEA~\cite{thavamani2021fovea} & 25.4 & 43.3 & 14.8 & 38.8 & 57.9 & 1.18 & 1.13 & 1.07 & \textcolor{black}{48.6}\\
    \cline{2-11}
    & LZU~\cite{Thavamani_2023_CVPR} & 24.9 & 42.8 & 14.1 & 38.5 & \textbf{58.5} & - & - & - & \textcolor{black}{49.0}\\
    \cline{2-11}
   & ZoomDet & \textbf{25.9} & \textbf{44.4} & \textbf{15.7} & \textbf{39.1} & 57.8 & \textbf{1.24} & \textbf{1.18} & 1.05  & \textcolor{black}{48.2}\\
    \cline{2-11}
   \hline
    \multirow{5}*{UAVDT} & Uniform & 19.6 & 33.3 & 13.1 & 32.2 & 26.7 & 1.00 & 1.00 & 1.00 & \textcolor{black}{43.7}\\
    \cline{2-11}
   & Saliency~\cite{recasens2018learning} & 20.7 & 34.0 & 13.4 & 33.5 & 28.1 & \textbf{1.52} & 1.09 & 1.02 & \textcolor{black}{48.1}\\
   \cline{2-11}
   & FOVEA~\cite{thavamani2021fovea} & 21.2 & 34.6 & 14.1 & 33.9 & 28.0 & 1.41 & 1.07 & 1.04 & \textcolor{black}{48.6}\\
   \cline{2-11}
   & LZU~\cite{Thavamani_2023_CVPR} & 20.8 & 35.0 & 13.0 & 32.8 & 28.7 & - & - & - & \textcolor{black}{49.0}\\
  \cline{2-11}
   & ZoomDet & \textbf{21.9} & \textbf{35.2} & \textbf{15.4} & \textbf{34.7} & \textbf{29.2}& 1.47 & \textbf{1.10} & \textbf{1.09} & \textcolor{black}{48.2}\\
      \hline
    \multirow{5}*{SeaDronesSee} & Uniform & 43.8 & 71.8 & 30.9 & \textbf{48.0} & 66.6 & 1.00 & 1.00 & 1.00 & \textcolor{black}{43.7}\\
    \cline{2-11}
   & Saliency~\cite{recasens2018learning} & 45.1 & 73.9 & 35.1 & 47.8 & 66.1 & 2.05 & 1.10 & 1.03 & \textcolor{black}{48.1}\\
    \cline{2-11}
   & FOVEA~\cite{thavamani2021fovea} & 44.9 & 74.5 & 35.3 & 47.0 & 65.7 & 2.01 & 1.22 & 1.05 & \textcolor{black}{48.6}\\
    \cline{2-11}
    & LZU~\cite{Thavamani_2023_CVPR} & 45.0 & 75.0 & 34.8 & 47.2 & 66.5 & - & - & - & \textcolor{black}{49.0}\\
    \cline{2-11}
   & ZoomDet & \textbf{46.2} & \textbf{75.0} & \textbf{38.9} & 47.4 & \textbf{66.8} & \textbf{2.90} & \textbf{1.32} & \textbf{1.09} & \textcolor{black}{48.2}\\
   \hline
\end{tabular}}
\label{tab:tab2}
\end{table*}

\subsection{Main Results}

\textcolor{black}{\subsubsection{Results Comparing to Non-uniform Zooming Methods}} The main experiment results on Faster R-CNN and YOLOv8-s are first presented. A zoom-in ratio metric (ZR) on the evaluation set is computed to evaluate how much the objects are enlarged. Specifically, for each ground-truth bounding box, the box area after zooming is divided by the original box area to obtain the ratio. As shown in Tab.~\ref{tab:tab1}, ZoomDet significantly outperforms the Faster R-CNN baseline with uniform downsampling on all three datasets. Remarkably, ZoomDet brings an 8.4 gain of mAP on SeaDronesSee dataset, especially for small objects (with a 26.2 absolute boost). Similar performance boost is observed with the YOLOv8 model, as in Tab.~\ref{tab:tab2}. For the ZR metric, the proposed method shows effective magnification capability, especially for small objects. For example, the small objects in SeaDroneSee are enlarged by over 2.6 times; we refer readers to qualitative results in the latter section.

The saliency-based parametrization in prior work~\cite{recasens2018learning,thavamani2021fovea} is also compared. Three comparison baselines are established: 1) \textbf{Saliency}: employing the 
saliency-based image warping in prior work~\cite{recasens2018learning}, and using the proposed box transformation for detector training and inference. 2) \textbf{FOVEA}: employing the FOVEA~\cite{thavamani2021fovea} framework, which adopts the saliency-based warping~\cite{recasens2018learning} and directly maps the predicted bounding boxes back to the original space during training. To compare with our method, the \textit{learned non-separable}~\cite{thavamani2021fovea} setting is used, which better predicts the small objects. 3) \textbf{LZU}: employing the LZU framework~\cite{Thavamani_2023_CVPR}, which develops saliency-based warping to extract image features on the warped space and then maps the features back to the original space.
In comparison to these approaches, the proposed method offers greater improvement to the uniform baseline. The object magnification ratios are also larger for different object scales, i.e., the $ZR$ metrics. 
The results demonstrate the superiority of the proposed framework, which delivers a better mapping parametrization and a flexible box transformation as discussed in~\ref{method_overview}. Meanwhile, the indirect learning paradigms of FOVEA~\cite{thavamani2021fovea} and LZU~\cite{Thavamani_2023_CVPR} hinder effective object detection training in the zoomed space, as they either map the bounding box predictions back to the original space during detector learning.

On the other hand, less improvement or some performance degradation is observed on larger objects, i.e., a drop in $AP_l$ and $AP_m$ with VisDrone and SeaDroneSee datasets, respectively. 
This is mainly due to the fact that the large objects are already large enough, while the magnification could lead to disturbance to the recognition accuracy of these objects.

\textcolor{black}{
\begin{table*}[]
\centering
\renewcommand{\arraystretch}{1.4}
\caption{\textcolor{black}{Comparison to other SOTA methods related to zoom-in technique. The result is obtained with VisDrone dataset. Pat.Z., Uni.Z., and Imp.Z. denote patch-based zooming, uniform zooming, and implicit zooming, respectively. F.P. denotes the computation cost in GFlops.}}
\setlength{\tabcolsep}{0.6mm}{\arrayrulecolor{black}
\begin{tabular}{>{\color{black}}c|>{\color{black}}c|>{\color{black}}c>{\color{black}}c>{\color{black}}c>{\color{black}}c>{\color{black}}c>{\color{black}}c|>{\color{black}}c|>{\color{black}}c>{\color{black}}c>{\color{black}}c>{\color{black}}c>{\color{black}}c>{\color{black}}c}
\hline
                                                & Methods & $AP$   & $AP_{50}$ & $AP_{s}$ & $AP_{m}$ & $AP_{l}$ & F.P. & Methods  & $AP$   & $AP_{50}$ & $AP_{s}$ & $AP_{m}$ & $AP_{l}$ & F.P.\\ \hline
\multirow{4}{*}{Pat.Z.} & Base  & 33.9 & 57.7   & 25.5  & 42.7  & 43.7 & 202.1  & Base  & 34.8 & 58.8   & 26.4  & 46.3  & 53.2  & 153.0    \\ \cline{2-15} 
                                                & ESOD\cite{liu2024esod}     & 36.0 & 59.7   & 28.3  & 45.9  & 45.0 & 135.4  & DG.\cite{xie2025density} & 37.4 & 61.7   & 30.2  & 47.3  & 52.0 & 305.9 \\ \cline{2-15} 
                                                & ZoomDet  & 35.1 & 60.2   & 28.1  & 44.7  & 44.3 & 210.2 & ZoomDet & 36.3 & 60.1   & 28.4  & 46.9  & 52.8 & 158.7 \\ \cline{2-15} 
                                                 & +Both    & 36.8 & 60.3   & 29.3  & 46.9  & 44.9 & 141.6  & +Both   & 38.1 & 62.6   & 31.4  & 48.6  & 53.5 & 329.4 \\ \hline
\multirow{4}{*}{Uni.Z.}                   & Base    & 25.0 & 43.7   & 13.9  & 39.0  & 56.7 & 123.1 & Base     & 25.5 & 44.9   & 14.4  & 39.9  & 57.2 & 16.4 \\ \cline{2-15} 
                                                & SR4I.\cite{kim2024beyond}   & 25.8 & 44.2   & 15.5  & 39.7  & 57.1 & 126.6 & Fe.-S.\cite{liu2024small} & 27.0 & 47.8   & 16.6  & 41.8  & 59.8 &  73.7 \\ \cline{2-15} 
                                                & ZoomDet & 26.1 & 44.6   & 15.5  & 40.2  & 57.0 & 128.9 & ZoomDet  & 27.5 & 48.5   & 16.9  & 42.5  & 60.7 & 21.2 \\ \cline{2-15} 
                                                & +Both   & -    & -      & -     & -     & -   & -  & +Both    & -    & -      & -     & -     & -   & -  \\ \hline
\multirow{4}{*}{Imp.Z.}                  & Base    & 24.4 & 40.5   & 14.5  & 36.7  & 46.0 & 39.5 & Base     & 26.6 & 43.5   & 15.9  & 39.2  & 48.6 & 68.2 \\ \cline{2-15} 
                                                & Rem.\cite{li2025remdet}  & 28.2 & 46.1   & 18.2  & 41.7  & 51.0 & 34.4 & RAF.\cite{bi2025towards}    & 30.0 & 48.5   & 19.5  & 44.1  & 58.6 & 72.0 \\ \cline{2-15} 
                                                & ZoomDet & 26.8 & 43.8   & 17.0  & 39.9  & 50.2 & 45.3 & ZoomDet  & 28.9 & 47.0   & 18.6  & 43.0  & 57.1 & 74.1 \\ \cline{2-15} 
                                                & +Both   & 29.2 & 47.5   & 19.4  & 43.7  & 55.8 & 39.6 & +Both    & 30.9 & 49.3   & 21.0  & 45.3  & 57.9 & 77.8 \\ \hline
\end{tabular}\arrayrulecolor{black}
}
\label{tab:comparison_recent_methods}
\end{table*}
}

\textcolor{black}{\subsubsection{Results Comparing to Uniform Zooming Methods}}

\textcolor{black}{
To make a more comprehensive comparison to the non-uniform zoom-in methods as well as the other SOTA methods that follow the zoom-in approach, we divide the other recent zooming-related methods 
into patch-based zooming\cite{meethal2023cascaded,liu2024esod,li2024saccadedet,xie2025density}, uniform zooming \cite{liu2024small,nihal2024blurry,kim2024beyond,asif2025novel,zhang2023superyolo} and implicit feature zooming\cite{wang2025learning,bian2025feature,bi2025towards,li2025remdet,wang2025detail,duan2025focal} and compare to them.
To have a fair comparison, the base object detectors of these compared works are adopted. Since the ZoomDet framework can be combined with the implicit zooming method and patch-based zooming method to improve detection accuracy, we additionally show the result of augmenting the compared methods with ZoomDet. For a uniform zooming method, as the zooming objective cannot be easily combined with the reconstruction objectives, the comparison results are reported without the combined results.
}
\textcolor{black}{
As shown in Tab.\ref{tab:comparison_recent_methods}, the observation can be summarized as 1) for patch-based zooming and implicit zooming, ZoomDet brings less improvement than the compared methods. 
e.g., 1.2 and 1.5 improvement in $AP$ compared to 2.1 and 2.6 improvement in $AP$ for patch-based ESOD\cite{liu2024esod} and DG\cite{xie2025density}. This is because they focus on different zooming perspectives, one addressing global patch-level object scale variation and the other addressing local object-level object scale variation.
Thus, when combined, optimal performance gain can be achieved, e.g., further boosting the ESOD\cite{liu2024esod} and DG\cite{xie2025density} by 0.8 and 0.7 in $AP$, 1.0 and 1.2 in $AP_{s}$.
A similar trend of performance improvement pattern can be observed for the implicit zooming methods. The combination also works well as the implicit zooming methods focus on feature-level object signal enhancement while ZoomDet focuses on image-level object magnification.
2) for uniform zooming methods, ZoomDet offers higher improvement, e.g., 2.0 improvement in AP compared to 1.5 for the Fe.-S.\cite{liu2024small} method.
The two actually work at the same level of the whole input image. Uniform zooming mainly improves the detection by enhancing object visual attributes such as the discriminative parts and edges.  
While ZoomDet can better enhance the object information by directly magnifying their spatial sizes, the improvement is greater.
}

\textcolor{black}{\subsubsection{Results on Natural Scene Image Datasets}}
\begin{table*}[htb]
\centering
\caption{Results on autonomous driving dataset Argoverse-HD~\cite{chang2019argoverse,li2020towards} and Common object category dataset COCO~\cite{lin2014microsoft}.}
\setlength{\abovecaptionskip}{5pt}
\renewcommand{\arraystretch}{1.3}
\setlength{\tabcolsep}{1.3mm}{
\begin{tabular}{c|c|cccccc>{\color{black}}c>{\color{black}}c>{\color{black}}c>{\color{black}}c}
   \hline
    \textbf{Dataset} &\textbf{Method} & $AP$ & $AP_{50}$ & $AP_{75}$ &$AP_{s}$ & $AP_{m}$ & $AP_{l}$ & $ZR_{s}$ & $ZR_{m}$ & $ZR_{l}$ & FLOPs\\
   \hline
    \multirow{4}*{\textbf{Argo}} & Uniform & 22.6 & 38.7 & 21.7 & 3.7 & 22.1 & 53.1 & 1.00 & 1.00 & 1.00 & 218.6\\
   & FOVEA~\cite{thavamani2021fovea} & 24.9 & 40.3 & 25.3 &  7.1 &  \textbf{27.7} &  50.6 & 1.41 & 1.37 & 1.04 & 225.1\\ 
    & LZU~\cite{Thavamani_2023_CVPR} & 25.3 & 43.0 & 24.6 & 6.1 & 25.9 & 52.6 & - & - & - & 226.3\\ 
    & ZoomDet & \textbf{26.4} & \textbf{43.7}  & \textbf{25.4} & \textbf{7.9} & 27.5 & \textbf{53.5} & 1.82 & 1.31 & 1.02 & 224.3 \\
   \hline
     \multirow{4}*{\textbf{COCO}} & Uniform & 39.0 & 60.5 & 42.3& 21.8 & 43.0 & \textbf{51.8} & 1.00 & 1.00 & 1.00 & 228.0\\
   & FOVEA~\cite{thavamani2021fovea} & 38.6 & 59.4 & 41.8& 20.8 & 42.1 & 51.0 & 1.12 & 1.03 & 1.02 & 234.5 \\
    & LZU~\cite{Thavamani_2023_CVPR} & 39.2 & 60.9 & 42.3& 21.9 & 43.2 & \textbf{52.0} & - & - & - &237.4\\
    & ZoomDet & \textbf{39.5} & \textbf{61.1} & \textbf{42.8} & \textbf{22.8} & \textbf{43.1} & 51.4 & 1.15 & 1.01 & 1.02 & 223.5\\
    \hline
\end{tabular}}
\label{tab:common_datasets}
\end{table*}
In addition to the UAV image domain, further experiments are conducted on the common natural scene datasets to test the generalizability of ZoomDet. the common object category dataset COCO~\cite{lin2014microsoft} and autonomous driving dataset Argoverse-HD~\cite{chang2019argoverse,li2020towards} are employed.

Some prior works, such as FOVEA~\cite{thavamani2021fovea} and LZU~\cite{Thavamani_2023_CVPR}, adopt the saliency-based image transformation~\cite{recasens2018learning} and developed a non-uniform sampling approach for object detection on autonomous driving scenes. Following these works, RetinaNet is employed as the object detector, and the performance of the proposed method is reported on the Argoverse-HD dataset and the widely used COCO benchmark. The results are shown in Tab.~\ref{tab:common_datasets}. The proposed method outperforms them, with a 3.8 absolute improvement of $AP$ to the Uniform sampling baseline. 
Notably, they use static spatial prior on the central horizontal areas to magnify the street scene objects that are mostly distributed in these areas, or leverage the box prediction of the previous frame as a spatial prior, while this work learn universal image-based object zooming and achieves better performance.

In addition, experiments on the COCO benchmark are conducted, which is one of the most common natural object datasets (80 common object categories). As shown in Tab.~\ref{tab:common_datasets}, unlike the above results, since the objects in COCO images are generally much larger and closer to nearby objects (many with occlusion), the improvement of ZoomDet is limited, i.e., 0.5 improvement in $AP$. The observation also corresponds with the failure case analysis above, which shows ZoomDet could lead to degenerated performance on large objects and crowded scenes. Notably, the proposed method brings noticeable performance improvement for small objects, i.e., 1.0 absolute gain in $AP_s$, which is the best among all comparing methods.

\textcolor{black}{\subsubsection{Results on Remote Sensing Image Dataset}}
\begin{table*}[]
\centering
\textcolor{black}{\caption{Results on DIOR dataset\cite{li2020object}. Same settings for Faster R-CNN and YOLOv8 as above are used.}}
\renewcommand{\arraystretch}{1.3}
\setlength{\tabcolsep}{1.6mm}{\arrayrulecolor{black}
\begin{tabular}{>{\color{black}}c|>{\color{black}}c|>{\color{black}}c>{\color{black}}c>{\color{black}}c>{\color{black}}c>{\color{black}}c>{\color{black}}c>{\color{black}}c>{\color{black}}c>{\color{black}}c}
\hline
Model                         &   -  &   $AP$ & $AP_{50}$ & $AP_{s}$ & $AP_{m}$ & $AP_{l}$ & $ZR_{s}$ & $ZR_{m}$ & $ZR_{l}$ & FLOPs \\ \hline
\multirow{2}{*}{Faster R-CNN} & Baseline &  21.2 & 50.1 & 7.0 & 26.8 & 54.4 & 1.00 &  1.00 & 1.00 & 176.5 \\ \cline{2-11} 
                              & ZoomDet  & 22.8 & 52.7 & 8.9 & 27.5 & 54.3 & 1.42 & 1.12 & 1.13 & 181.3\\ \hline
\multirow{2}{*}{YOLOv8}       & Baseline & 23.0 & 53.8 & 7.9 & 29.2 & 56.5 & 1.00 & 1.00 & 1.00 & 43.7 \\ \cline{2-11} 
                              & ZoomDet  & 24.3 & 55.9 & 9.4 & 29.9 & 56.8 & 1.30 & 1.14 & 1.05 & 48.2 \\ \hline
\end{tabular}\arrayrulecolor{black}
}
\label{tab:remote_sensing_data}
\end{table*}
\textcolor{black}{
While this work focuses on UAV object detection, as the paper title denotes, the method is general. Here, more experiments are conducted by adding the results to a remote sensing small object detection dataset, DIOR\cite{li2020object}. The dataset is a large-scale aerial object detection dataset with large-range object size variation and diverse visual qualities. The images are in resolution
800x800, and the total number of classes is 20. The images are divided into 5862 train images, 5863 validation images, and 11725 test images. The combination of train and validation sets is used to train the model and evaluate the result on the test set. The baseline model is the same one used in the main experiment, i.e., the Faster R-CNN and YOLOv8 model; the training setting and hyperparameters are the same.
}
\textcolor{black}{
As shown in Tab. 5, while the improvement is less than that in UAV captured detection datasets such as SeaDroneSee and UAVDT, it still provides an effective boost for detection, especially on the small objects. For example, Faster R-CNN is improved by 1.6 in $AP$ and 1.9 in $AP_{s}$; YOLO is improved by 1.3 in $AP$ and 1.5 in $AP_{s}$. Most likely, the background is more confusing for the remote sensing dataset, leading to less effective recognition for the offset prediction network, thus less magnification on the small objects. This is also a good point to conduct future work.
}

\textcolor{black}{\subsection{Empirical Analysis}}

\subsubsection{Ablation Study of Loss Hyperparameters}
\begin{figure}[h!]
	\centering
	\includegraphics[width=0.9\linewidth]{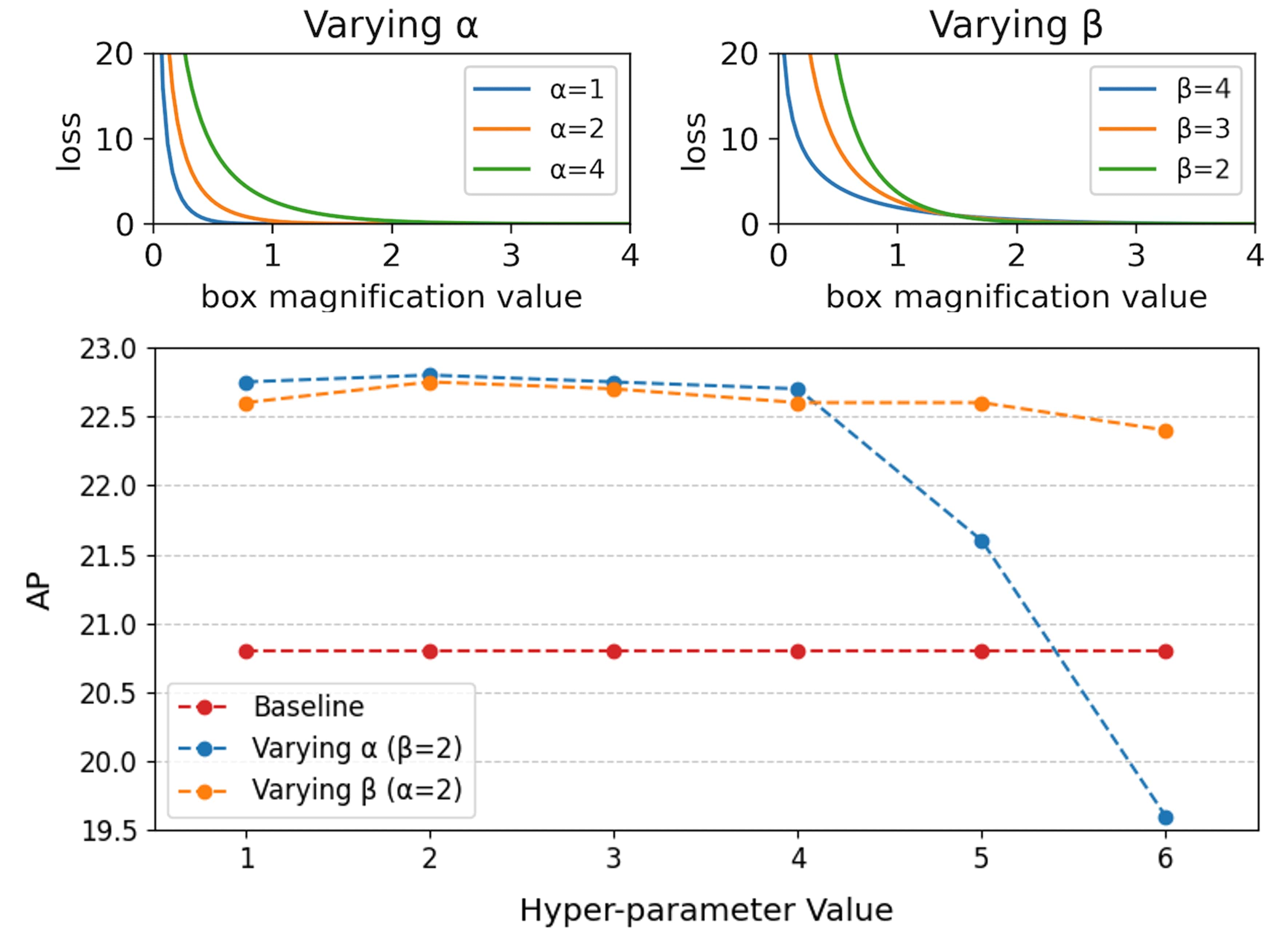}
	\caption{Ablation study on hyper-parameters $\alpha$ and $\beta$. The top two figures show the effect of these parameters on the zooming loss shape. The bottom figure shows the ablation results by independently varying each parameter with the other fixed. The results are obtained on VisDrone dataset.} 
	\label{supp:fig:ablation}
\end{figure}
As shown in Fig.~\ref{supp:fig:ablation}, a comprehensive ablation study is conducted on the hyperparameters $\alpha$ and $\beta$, by independently varying these parameters. 
Firstly, how they affect the loss shape is examined. As shown in the top two subfigures of Fig.~\ref{supp:fig:ablation}, it can be observed that larger $\alpha$ leads to a larger zero-loss threshold, while larger $\beta$ results in a sharper loss curve. These observations align well with the discussion in Sec.~\ref{method_obj_zooming_loss}.
Then, how the hyperparameters affect the performance directly is examined. As shown in the bottom subfigure of Fig.~\ref{supp:fig:ablation}, one is varying while the other is fixed. For $\alpha$, the performance stays at roughly 22.7 between the value range of 1-4, and it peaks at $\alpha=2$, while dropping sharply between 4-6. This is likely due to that too large $\alpha$ forces the magnification to collapse and results in extreme distortion. The performance improvement is rather stable for $\beta$, with more than 1.7 absolute improvement in AP (i.e., from 20.8 for baseline to 22.5) between 1-6, and peaks at $\beta=2$.
In summary, although fluctuations exist, good performance improvement can be achieved with a wide value range of these hyperparameters.

\textcolor{black}{\subsubsection{Ablation Study of Offset Prediction Network}}
\begin{table*}[h!]
\centering
\textcolor{black}{\caption{Ablation study on the network depth, architecture, and offset map resolution. The result is obtained on SeaDroneSee dataset.}}
\setlength{\tabcolsep}{1.6mm}{\arrayrulecolor{black}
\begin{tabular}{>{\color{black}}c>{\color{black}}c>{\color{black}}c|>{\color{black}}c>{\color{black}}c>{\color{black}}c|>{\color{black}}c>{\color{black}}c>{\color{black}}c}
\hline
Depth     & $AP$ & $AP_{s}$ & Architecture & $AP$ & $AP_{s}$ & Resolution & $AP$ & $AP_{s}$ \\ \hline
ResNet18  & 46.2  & 38.9  & ResNet50     &  46.5  & 39.3  & 8x         &  46.2 & 38.9  \\ \hline
ResNet34  & 46.5  & 39.4  & ConvNeXt-T   &  46.3  &  39.0  & 4x         &  46.9  &  40.9 \\ \hline
ResNet50  & 46.5 & 39.3  & Swin-T       &  45.8  &  37.5  & 2x         & 46.4  & 39.4  \\ \hline
ResNet101 & 46.0  & 38.2  &DeiT-S       &  45.5 & 37.2  & 1x         &  46.5  & 39.3  \\ \hline
\end{tabular}\arrayrulecolor{black}
}
\label{ablation_depth_arch_res}
\end{table*}

\textcolor{black}{To further examine the effect of offset prediction resolution, offset network architecture, and depth, an additional ablation study is conducted.
For network depth, ResNet with 18, 34, 50, and 101 layers are compared; For architecture, popular architectures across convolution networks and transformers are compared. These include ResNet\cite{he2016deep}, ConvNeXT\cite{liu2022convnet}, Swin Transformer\cite{liu2021swin}, DeiT\cite{touvron2021training}.
The architectures with a similar number of parameters (about 20M) and computation cost (about 25GFlops for the ImageNet full classification model) are chosen. For resolution, U-net\cite{ronneberger2015u} is employed to up-sample the default 8x resolution to 4x, 2x, and 1x prediction.}

\textcolor{black}{As shown in the Tab.\ref{ablation_depth_arch_res}, offset networks with varying depth, architecture, and output resolution are examined.
1) depth: for ResNet with varying depth, the performance does not vary much, with 0.3 AP improvement on deeper ResNet34 and ResNet50, while declining by 0.2 AP on the ResNet101. The result indicates that the offset prediction is a low-level task, which does not benefit much from deeper networks. 2) architecture: the convolution networks are generally better than the transformer-based backbone architecture, i.e., average performance is 46.4 and 45.7 for them, respectively. The result shows that convolution-based offset prediction may be better at capturing low-level object information.
3) resolution: increasing the resolution could bring improvements, e.g., lifting the 8x output resolution to 4x improves the AP by 0.7.
However, the gain diminishes at 2x, with only 0.2 improvement. Given 
finer resolution could introduce a large amount of computation,
The result means that the default 8x resolution could be a good "sweet spot" in the trade-off. \textit{Yet we would like to note that there could be improvement on the current learning paradigm of the offset prediction that leads to much greater improvement with higher resolution, e.g., sophisticated loss objectives that consider the context of objects, or the statistics of object density on nearby regions.} These are left for a future study.}

\textcolor{black}{\subsubsection{Error Analysis of Box Transformation}}
\begin{figure}[h!]
	\centering
	\includegraphics[width=0.9\linewidth]{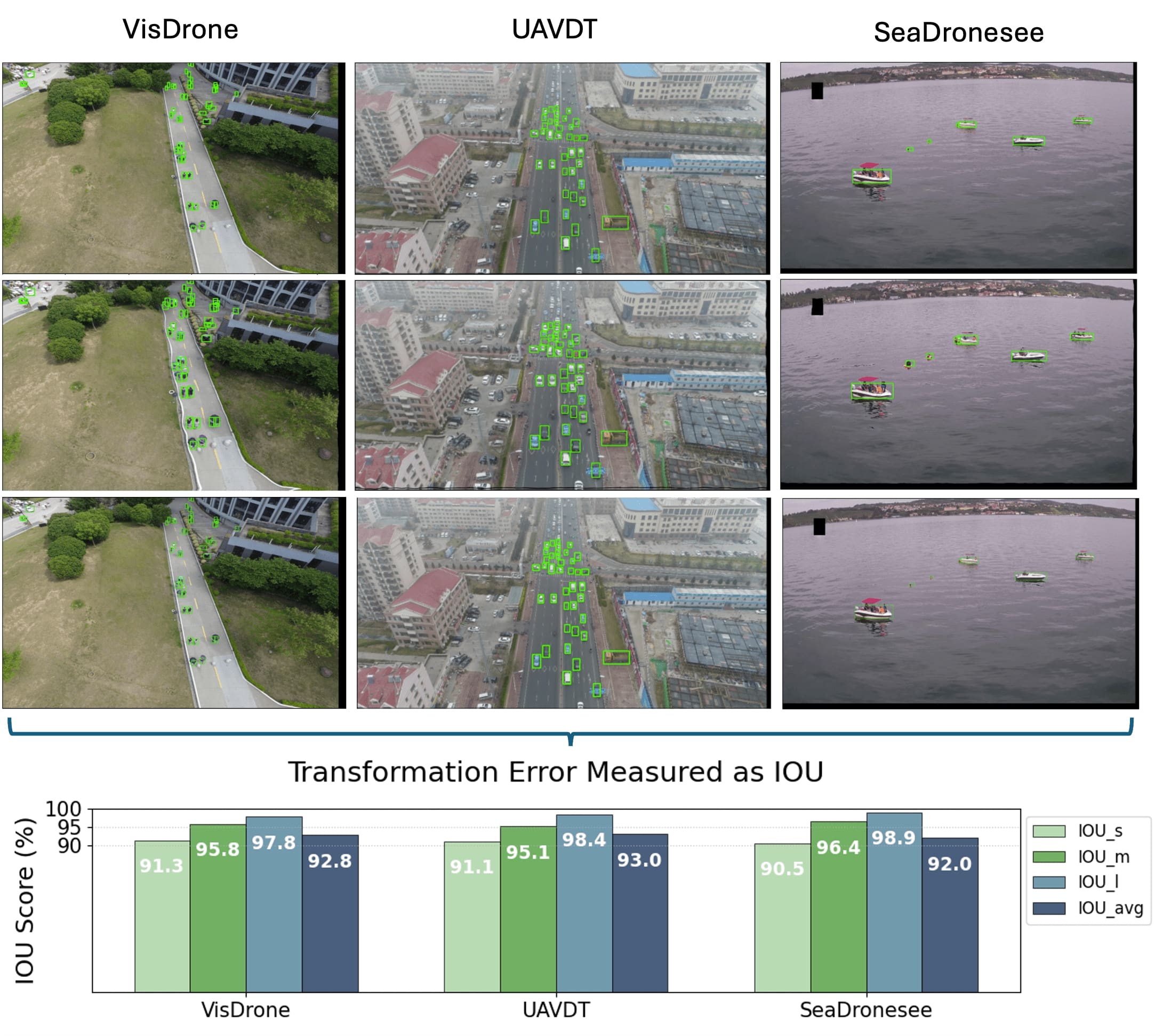}
	\caption{Empirical validation of error induced with forward and backward bounding box transformation. \textbf{The top row} shows example GT bounding boxes on the original image, \textbf{the second row} shows the transformed GT bounding boxes on the zoomed image, \textbf{the last row} shows the GT bounding boxes transformed back to the original space. The bar chart shows the IoU between the original and the backward-transformed boxes.}
	\label{fig:IOU}
\end{figure}
As discussed in the main paper, the errors induced by the proposed bounding box label transformation scheme are calculated.
Given the learned transformation, ground-truth bounding boxes are transformed to the zoomed image space and transformed back to the original space to compute the IoU with the original bounding box. As shown in the Fig.~\ref{fig:IOU}, in all three UAV image datasets, the proposed label transformation strategy achieves more than 92\% IoU on average with the original labels. Meanwhile, the box label transferred to the zoomed space still bounds the target object well and thus maintains good consistency of the supervisory signals.

\textcolor{black}{\subsubsection{Cost Analysis in Memory and Computation}}
Additionally, Tab.~\ref{tab:tab5} shows the extra cost in parameters, computation, and latency brought by ZoomDet, the ZoomDet-rb is a reduced version where the backbone is reused for OffsetNet backbone. Specifically, the stand-alone OffsetNet is removed to save computation. The offsets are instead predicted through the object detector backbone model by appending a single-layer convolution. Thus, the backbone is reused as OffsetNet's backbone and trained together. The zooming of objects is now conducted with image features instead of raw images. With ZoomDet-rb, the performance is slightly lower than standard ZoomDet, but the extra cost in parameters, computation, and latency is significantly reduced, as the additional cost is with only a single convolution layer and the transformations.

\begin{table*}[h!]
\centering
\caption{Extra cost of ZoomDet, including parameters, Flops, and latency. ZoomDet-rb denotes the proposed backbone re-use version that reuses the detector backbone for OffsetNet backbone. Results obtained with SeaDronesSee dataset.}
\renewcommand{\arraystretch}{1.1}
\setlength{\tabcolsep}{1.0mm}{
\begin{tabular}{c|c|cccc}
\hline
Model                   & Method     & AP   & Params & FLOPs & Latency \\ \hline
\multirow{3}{*}{FRCNN}  & uniform    & 34.9 & 45.8M      & 157.4G    & 41.5ms        \\
                        & ZoomDet    & 43.3 & 46.5M     & 163.3G    & 44.7ms        \\
                        & ZoomDet-rb & 42.0 & 45.8M      & 157.5G    & 41.6ms        \\\hline
\multirow{3}{*}{YOLOv8} & uniform    & 43.8 & 11.5M      & 43.7G     & 12.3ms        \\
                        & ZoomDet    & 46.2 & 12.2M      & 48.2G     & 14.0ms        \\
                        & ZoomDet-rb & 45.9 & 11.5M      & 43.8G     & 12.5ms       \\\hline
\end{tabular}}
\label{tab:tab5}
\end{table*}

\begin{figure*}[h!]
	\centering
	\includegraphics[width=\linewidth]{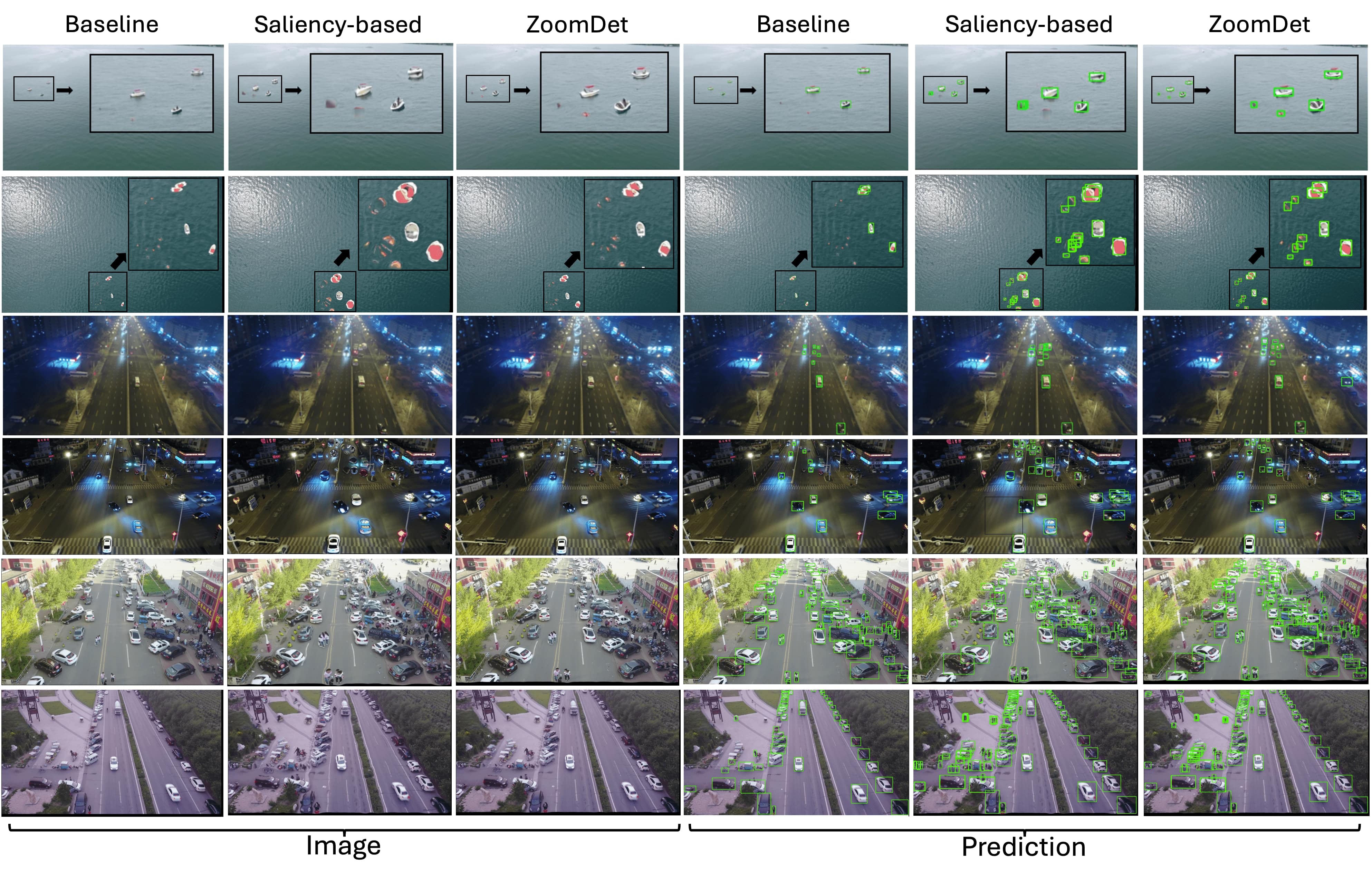}
	\caption{\textbf{Example of prediction results on SeaDroneSee(top two rows), UAVDT(middle two rows) and VisDrone(bottom two rows) datasets}. The proposed method helps the model better detect some small objects that are ignored by the baseline. And the results are better than those of the saliency-based baseline discussed in the method section. \textit{Best viewed with zoom-in.}} 
	\label{fig:prediction}
\end{figure*}

\begin{figure*}[h!]
	\centering
	\includegraphics[width=\linewidth]{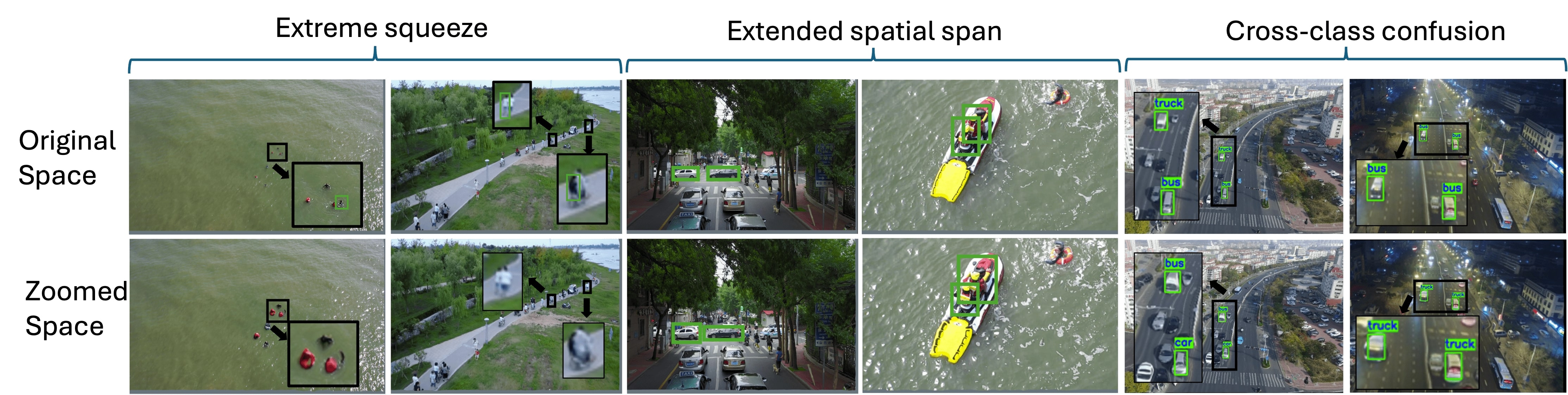}
	\caption{Failure case analysis. Three main failure cases are identified, which are caused by the distortion introduced by ZoomDet. \textbf{Extreme squeeze}: The objects in the green bounding boxes are missed as they are extremely squeezed in the zoomed space and not recognized. \textbf{Extended spatial span}: Compared to the ground-truth bounding boxes in the top row, the predictions are spatially expanded with excessive spatial span. \textbf{Cross-class confusion}: Compared to the GT classes in the top row, the predictions in the bottom are wrong as object magnification is not uniform and thus leads to cross-class confusion between similar UAV object classes (for clarity, the other objects are skipped in visualization). Best viewed with Zoom-in.} 
	\label{supp:fig:failurecase}
\end{figure*}

\textcolor{black}{\subsection{Discussion}}
\textcolor{black}{\subsubsection{Prediction Visualization and Failure Cases}}
We visualize the zoomed image and detection results in the zoomed space. These images and results are compared to the baseline without any zooming and the saliency-based zooming parametrization as discussed in the method section.
As shown in Fig.~\ref{fig:prediction}, the proposed method helps detect small objects that are missed by the baseline, such as the \textit{swimmer} objects in the top two row images from Seadronesee dataset. While the saliency-based one can cause heavier distortion to the objects, and the detection results are worse, \textit{e.g.}, missed and duplicated detections. 
The observation also holds for the sample images from UAVDT dataset and VisDrone dataset.



ZoomDet magnifies on the objects and thus helps improve detection performance; however, the ``Zoom-in" could sometimes fail and lead to degenerated results. As shown in Fig.~\ref{supp:fig:failurecase}, the three major types of failure causes are summarized, including i) Extreme squeeze of objects. The image transformation could sometimes cause extreme magnification of certain objects, which leads to unexpected squeezing of nearby objects. These squeezed objects may be missed by the subsequent object detector. ii) Extended spatial span. The object magnification could lead to expanded prediction of bounding boxes while the actual spatial span is smaller. As shown in the fourth column of Fig.~\ref{supp:fig:failurecase}, the bounding box of one \textit{swimmer} instance is much larger than the GT box. iii) Cross-class confusion. The object magnification could cause confusion between some object classes. For example, after image transformation, the small object instances of \textit{bus} and \textit{truck} classes could be more similar and thus misclassified. As shown in the right-most column of Fig.~\ref{supp:fig:failurecase}, the two \textit{bus} instances are misclassified as a truck with ZoomDet.

\subsubsection{Comparison of Bounding Box Distribution}
\begin{figure*}[h]
	\centering
	\includegraphics[width=\linewidth]{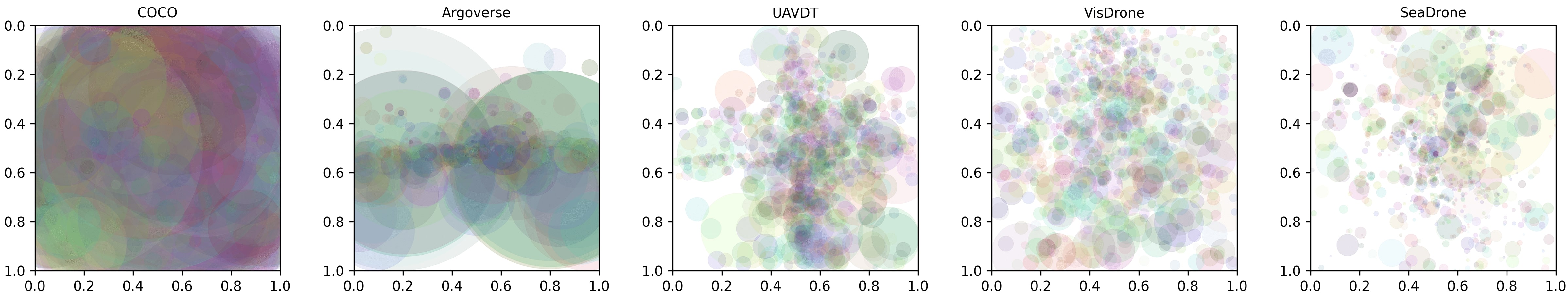}
	\caption{Comparison of bounding box distribution between datasets. The figure shows the spatial location and size distribution of bounding box distributions on the five datasets:
    COCO (a common object detection benchmark), Argoverse (an autonomous driving scenario) used in prior methods, and the three UAV object detection datasets used in this work, i.e., UAVDT, VisDrone, SeaDronesee. 1000 normalized bounding boxes are randomly sampled and used to plot the distribution of object bounding boxes as circles, of which the radius is the square root of the bounding box area.}
    \label{fig:box_dist_comparison}
\end{figure*}
As discussed in the main results, ZoomDet introduces stronger performance improvement in the three UAV image datasets (i.e., Seadronesee, UAVDT, VisDrone) and the autonomous driving dataset (i.e., Argoverse), but offers minor improvement on the common object detection dataset (i.e., COCO). This is likely caused by the different object distribution in these datasets. To verify the assumption, the bounding box distribution of these three datasets is further examined. The distribution of the normalized bounding box is plotted. The distribution is plotted as circles on normalized image coordinates. The half square root of the bounding box area is plotted as the radius of each circle, which helps show the size distribution of bounding boxes. As shown in Fig.~\ref{fig:box_dist_comparison}, COCO dataset contains a lot of large objects, while the three UAV image datasets contain many more small objects. Especially for SeaDronesee dataset, there are a lot of extremely small objects, which explains the higher performance boost on this dataset. The Argoverse dataset also contains more large objects than that of the three UAV image datasets, but it also contains a lot of small objects. Most apparently, the objects are mainly distributed in the central horizontal image area due to the viewing angle of autonomous driving. As a result, ZoomDet could trivially learn to magnify the central horizontal area to boost the detection performance. This explains the better performance improvement compared to that of the COCO dataset.



\textcolor{black}{\subsubsection{Application to Vision-language Model}}
\label{app:geochat}
\begin{figure*}[th!]
	\centering
	\includegraphics[width=\linewidth]{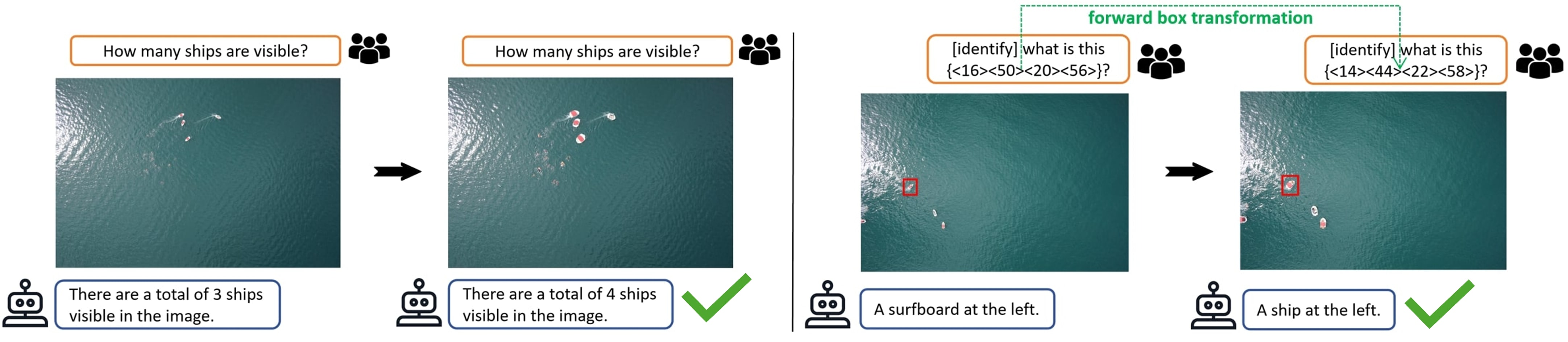}
	\caption{Examples with GeoChat~\cite{kuckreja2023geochat} for visual Q\&A on UAV images. ZoomDet image transformation and bounding box transformation are integrated into the GeoChat model. The proposed zoomed-in images help large aerial visual language models to better perceive objects in the images.}
	\label{fig:geochat}
\end{figure*}
Since our method processes the image and obtains a zoomed version with objects magnified, it can be employed for other scenarios where image-based recognition is performed. \textit{Hence, we explore applying ZoomDet images and bounding box transformation to a recent large aerial vision-language model named \textbf{GeoChat}~\cite{kuckreja2023geochat}, for the task of visual question answering and region-based captioning.} The qualitative result shows potential for further extending the proposed method to these scenarios. 

Here, some visual examples are presented.
As shown in Fig.\ref{fig:geochat},
The visual Q\&A results get improved, specifically, i) when tasked to count the number of ships, the proposed zoomed-in image magnifies the meaningful foreground area and thus helps the model perceive the objects; ii) when tasked to identify the object in the given bounding box area, the user-given bounding box is transformed with the proposed box transformation, and feed the new image as well as the transformed bounding box to the model. Similarly, the ship object is correctly recognized, which may indicate that zooming helps enhance the category discriminative information.

\section{Conclusion}
This work introduces a ``zoom and detect" framework for object detection on UAV images, where objects are generally small and sparsely distributed.
The main motivation is to zoom in non-uniformly on the object regions and thus help the following object detectors to better recognize them. 
To achieve an efficient zooming procedure and enable object detector learning on the zoomed image space, an offset prediction mechanism is developed to regress the non-uniform image sampling grid.  Based on the offset-based zooming, we further introduce a box transformation method to help transform the bounding boxes during training and inference. 
ZoomDet shows significant performance gains on several UAV object detection datasets and incurs a minor extra cost.

\textbf{Limitations and Future Work}. While being efficient, ZoomDet currently only supports object detection; the other visual object perception tasks, like instance segmentation and semantic segmentation, are not supported in the current label transformation scheme. It is thus important to develop an efficient transformation method for other tasks. Furthermore, it is observed that the training procedure of ZoomDet can be unstable occasionally, especially on crowded scene datasets; this is likely due to the competition of neighboring objects in the object zooming loss optimization. Methodologies like multi-objects combined zooming objective will be explored to improve this aspect.

\paragraph{Acknowledgements}

This work is supported by the National Science Foundation of China under Grant 62506249,
Natural Science Foundation of Sichuan under grant 2024NSFSC1462, the Fundamental Research Funds for the Central Universities under grant YJ202342, Natural Science Foundation of Sichuan under Grant 24NSFSC3404, the National Major Scientific Instruments and Equipments Development Project of National Natural Science Foundation of China under Grant 62427820, and the Science Fund for Creative Research Groups of Sichuan Province Natural Science Foundation under Grant 2024NSFTD0035.




\bibliographystyle{elsarticle-num} 
\bibliography{new-cas-refs.bib}

\end{document}